\documentclass{ieeeaccess}

\usepackage{soul}
\usepackage{algorithm}
\usepackage{listings}
\usepackage{multirow}
\usepackage{lineno}
\usepackage{hyperref}
\usepackage{algpseudocode}
\usepackage{booktabs}
\usepackage{mathtools}
\soulregister\cite7
\soulregister\ref7
\soulregister\sc7
\usepackage{eurosym}
\usepackage{siunitx}
\usepackage{textgreek}
\usepackage{tablefootnote}
\renewcommand{\theenumi}{\Alph{enumi}}
\usepackage{pifont}
\newcommand{\cmark}{\ding{51}}%
\newcommand{\xmark}{\ding{55}}
\usepackage{subfig}
\usepackage{amsmath,amssymb}
\usepackage[table]{xcolor}
\def\textsc#1{\textnormal{{\sc #1}}}
\usepackage{caption}
\DeclareCaptionFont{ieeeblue}{\color{accessblue}}
\DeclareCaptionLabelFormat{myformat}{\raggedright\figcapfont{\textbf{#1}\textbf{#2}}}
\def\BibTeX{{\rm B\kern-.05em{\sc i\kern-.025em b}\kern-.08em
		T\kern-.1667em\lower.7ex\hbox{E}\kern-.125emX}}

\newcommand{\expnumber}[2]{{#1}\text{e}\textsuperscript{#2}}
\newcommand{\sctxt}[1]{{\sc #1}}
\newcommand{\red}[1]{{\color{red} #1}}
\soulregister{\sctxt}{7}
\soulregister{\red}{7}
\soulregister{\expnumber}{7}
\soulregister{\SI}{7}
\soulregister{\percent}{7}

\usepackage{blindtext}
\usepackage{hyperref}
\usepackage{nameref}

\newcounter{mylabelcounter}

\makeatletter
\newcommand{\labelText}[2]{%
	\refstepcounter{mylabelcounter}%
	\immediate\write\@auxout{%
		\string\newlabel{#2}{{\unexpanded{#1}}{\thepage}{{\unexpanded{#1}}}{mylabelcounter.\number\value{mylabelcounter}}{}}%
	}
}
\makeatother

\makeatletter
\newcommand\footnoteref[1]{\protected@xdef\@thefnmark{\ref{#1}}\@footnotemark}
\makeatother

\begin{document}
	
	\history{Date of publication xxxx 00, 0000, date of current version xxxx 00, 0000.}
	\doi{xx.xxxx/ACCESS.xxxx.DOI}
	
	\title{Unraveling Emotions with Pre-Trained Models}
	
	\author{\uppercase{Alejandro Pajón-Sanmartín}\authorrefmark{1}, \uppercase{Francisco de Arriba-Pérez}\authorrefmark{1}, \uppercase{Silvia García-Méndez}\authorrefmark{1}, \uppercase{Fátima Leal}\authorrefmark{2}, \uppercase{Benedita Malheiro}\authorrefmark{3}, and \uppercase{Juan Carlos Burguillo-Rial}\authorrefmark{1}}
	
	\address[1]{Information Technologies Group, atlanTTic, University of Vigo, Vigo, Spain}
	
	\address[2]{Research on Economics, Management and Information Technologies, Universidade Portucalense, Porto, Portugal}
	
	\address[3]{ISEP, Polytechnic of Porto, Porto, Portugal \& Institute for Systems and Computer Engineering, Technology and Science, Porto, Portugal}
	
	\tfootnote{This work was partially supported by Portuguese national funds through FCT -- Fundação para a Ciência e a Tecnologia (Portuguese Foundation for Science and Technology) -- as part of project UIDP/50014/2020 (DOI: 10.54499/UIDP/50014/2020 | \small{\url{https://doi.org/10.54499/UIDP/50014/2020}}).}
	
	\markboth
	{Alejandro Pajón-Sanmartín \headeretal: Unraveling Emotions with Pre-Trained Models}
	{Alejandro Pajón-Sanmartín \headeretal: Unraveling Emotions with Pre-Trained Models}
	
	\corresp{Corresponding author: Silvia García-Méndez (e-mail: sgarcia@gti.uvigo.es).}
	
	\begin{abstract}
		Transformer models have significantly advanced the field of emotion recognition. However, there are still open challenges when exploring open-ended queries for Large Language Models (\textsc{llm}s). Although current models offer good results, automatic emotion analysis in open texts presents significant challenges, such as contextual ambiguity, linguistic variability, and difficulty interpreting complex emotional expressions. These limitations make the direct application of generalist models difficult. Accordingly, this work compares the effectiveness of fine-tuning and prompt engineering in emotion detection in three distinct scenarios: (\textit{i}) performance of fine-tuned pre-trained models and general-purpose \textsc{llm}s using simple prompts; (\textit{ii}) effectiveness of different emotion prompt designs with \textsc{llm}s; and (\textit{iii}) impact of emotion grouping techniques on these models. Experimental tests attain metrics above \SI{70}{\percent} with a fine-tuned pre-trained model for emotion recognition. Moreover, the findings highlight that \textsc{llm}s require structured prompt engineering and emotion grouping to enhance their performance. These advancements improve sentiment analysis, human-computer interaction, and understanding of user behavior across various domains.
	\end{abstract}
	
	\begin{keywords}Emotion recognition, large language models, natural language processing, open-ended responses, prompt engineering, transformer models.
	\end{keywords}
	
	%\linenumbers
	
	\titlepgskip=-15pt
	
	\maketitle
	
	\section{Introduction}
	
	Emotion recognition is a task for Natural Language Processing (\textsc{nlp}), which enables machines to understand and respond to human emotions embedded in text. Emotion recognition analyzes texts to detect and classify emotions such as sadness, joy, love, anger, fear, and surprise \cite{alswaidan2020survey}. This ability is essential for various applications, including sentiment analysis, customer service, mental health monitoring, and human-computer interaction. The advent of transformer models has significantly advanced the field, offering good accuracy in capturing human emotions \cite{khan2025memocmt}. \textsc{nlp} models have gained significant importance in recent years, primarily due to advances in their ability to analyze and understand human language in an automated manner \cite{khurana2023natural}. These improvements have been significantly enhanced, and the technology has been popularized by companies such as OpenAI and Google.
	
	Transformer models, introduced by \cite{vaswani2017attention}, are a type of deep learning architecture that leverages self-attention mechanisms to process data sequences. These models can capture long-range dependencies and contextual information more effectively than previous architectures, such as Recurrent Neural Networks (\textsc{rnn}s) and their subset, Long Short-term Memory Networks (\textsc{lstm}s) \cite{zheng2024modeling}. Building on transformer architecture, Large Language Models (\textsc{llm}s) are deep learning models trained on vast amounts of text data to understand and generate human language. \textsc{llm}s, such as the Generative Pre-trained Transformer models (\textit{e.g.}, \textsc{gpt}-3  and \textsc{gpt}-4), leverage the extensive knowledge gained during pre-training on diverse text corpora and can be fine-tuned for specific tasks, achieving high accuracy in various \textsc{nlp} applications. In contrast, traditional machine learning (\textsc{ml}) models, such as logistic regression, support vector machines, and decision trees, typically rely on manually engineered features and are trained on specific tasks with limited data sets. While effective for certain applications, these models often lack the scalability and contextual understanding of transformer-based architectures and \textsc{llm}s.
	
	Transformer models, such as Bidirectional Encoder Representations from Transformers (\textsc{bert}) \textsc{gpt}, have demonstrated superior performance in a wide range of \textsc{nlp} tasks due to their ability to capture long-range dependencies and contextual information effectively. These models can be employed for prompt engineering and fine-tuning. Prompt engineering formulates specific prompts to guide the model's responses without altering internal parameters. At the same time, fine-tuning adjusts the model's weights by training it on a task-specific dataset.
	
	One of the most challenging approaches to emotion recognition is the joint exploration of open-ended queries and transformer models. Unlike structured data, open-ended questions are rich in context and detail, providing a comprehensive view of the inquirer's emotions and desires. This complexity presents significant challenges for emotion detection models, as they must accurately interpret diverse expressions and contexts of emotion. To address this challenge, this work compares the effectiveness of prompt engineering and fine-tuning in emotion detection with open-ended questions. 
	
	This paper analyses three distinct scenarios in the context of emotion recognition: (\textit{i}) performance of fine-tuned pre-trained models and general-purpose \textsc{llm}s using simple prompts; (\textit{ii}) effectiveness of different emotion prompt designs; and (\textit{iii}) impact of different emotion grouping techniques on the performance of \textsc{llm}s. Regarding emotion recognition with open-ended responses, the goal is to identify the most effective methods to improve the accuracy and reliability of \textsc{bert}, \textsc{R}o\textsc{bert}a, Gemma, \textsc{gpt}-3.5, and \textsc{ll}a\textsc{ma}-3. The results with six emotion classes show that fine-tuned \textsc{bert} and spaCy models are effective at emotion detection with at least \SI{80}{\percent} accuracy, while general-purpose \textsc{llm}s using prompt engineering achieve only around \SI{50}{\percent} accuracy. The latter models reach \SI{80}{\percent} accuracy when focusing solely on two emotion categories.
	
	This document is structured as follows. Section \ref{sec:related_work} surveys multiple works based on traditional and transformer-based models. Section \ref{sec:proposed_method} details the proposed method, describing the explored scenarios. Moreover, Section \ref{sec:experimental_results} presents and discusses the results compared to competing works from the literature. Finally, Section \ref{sec:conclusion} makes the final remarks.
	
	Despite recent advances, automatic emotion recognition in open-ended responses still presents significant research gaps. In particular, general-purpose \textsc{llm}s show limitations in handling complex emotional expressions without specific tuning, and there is a critical dependence on prompt design. Furthermore, systematic comparisons between fine-tuning-based approaches and prompt engineering remain scarce in open-ended natural language contexts. Given these opportunities for contribution, the following objectives are proposed:
	
	\begin{itemize}
		\item To comparatively evaluate the performance of fine-tuned pre-trained models and general-purpose \textsc{llm}s using prompt engineering in emotion recognition tasks.
		
		\item To analyze the impact of different types of prompts on model performance.
		
		\item To study how emotion grouping affects the emotion detection capabilities of \textsc{llm}s.
	\end{itemize}
	
	\section{Related work}
	\label{sec:related_work}
	
	Open-ended questions are a qualitative data collection method in which respondents can answer questions in their own words rather than selecting from predetermined options. Unlike closed-ended questions, which offer a fixed set of options (\textit{e.g.}, yes/no, multiple-choice), open-ended queries enable respondents to express their thoughts, feelings, and experiences. Additionally, open-ended questions are employed across various domains, including psychology \cite{obergassel2025adaptation}, sociology \cite{kosimov2025statistical}, marketing \cite{helm2024no}, and education \cite{olvet2025open}. Typically, they are employed in surveys, interviews, and focus groups to gather individual feedback and derive insights.
	
	The analysis of open-ended responses presents several challenges. The unstructured nature of the data requires sophisticated methods of processing and interpretation. Traditionally, researchers have relied on manual classification, where themes and patterns are identified through an intensive reading process and categorization of responses \cite{hansen2024integrating}. However, recent advancements in \textsc{nlp} and \textsc{ml} have automated the analysis of open-ended data, enabling efficient and scalable evaluation \cite{gweon2024automated}. \textsc{nlp} has grown significantly in recent years, primarily due to its flexibility to adapt to a wide range of applications and to generate complex responses with minimal instructions. 
	
	Additionally, in emotional evaluation, open-ended questions are a powerful source of information, as emotions are complex and multifaceted. In this context, open questions can help individuals understand how they feel and express emotions, which is fundamental to understanding consumer behavior \cite{vyas2024exploring}, inferring mental health states \cite{ajayi2024neural}, or improving human-computer interaction \cite{thiripurasundari2024speech}. To address this challenge, this research explores the ability of \textsc{nlp} models (\textsc{llm}s and traditional approaches) to perform text-based emotional evaluation, aiming to comprehend and analyze the emotions expressed in a text.
	
	The literature on emotional evaluation includes methodologies for analyzing emotional content \cite{park2025integrated}, \textsc{ml} classification methods \cite{ahamad2025exploring}, \textsc{nlp} techniques \cite{mishra2025unveiling}, and the implications of these findings for various domains \cite{houssein2025leveraging}. Understanding human emotions in texts enables the analysis of human communication, leading to more informed decisions, better services, and improved outcomes across various fields \cite{lerner2024emotions}. The integration of emotional evaluation into the analysis of open-ended responses represents a critical advancement in harnessing the full potential of textual data.
	
	\subsection{\textsc{ml} models for emotion recognition}
	
	\textsc{ml} models have revolutionized the field of emotional evaluation, offering tools for extracting and interpreting the emotional content embedded in textual data \cite{alslaity2024machine}. These models leverage advanced algorithms to analyze vast amounts of text efficiently and accurately, identifying patterns and emotional cues that may be imperceptible to traditional manual analysis.
	
	Supervised learning involves training models with labeled data sets where the emotional categories of text samples are predefined. This approach allows the model to learn the relationship between linguistic features and specific emotions. Standard supervised learning algorithms used in emotional evaluation include: (\textit{i}) Naive Bayes (\textsc{nb}); (\textit{ii}) Support Vector Machines (\textsc{svm}); (\textit{iii}) Neural Networks (\textsc{nn}); (\textit{iv}) Logistic Regression (\textsc{lr}); (\textit{v}) K-Nearest Neighbours (\textsc{knn}); and/or (\textit{vi}) Boosting and Gradient ensemble techniques, \textit{e.g.}, Random Forest. These \textsc{ml} algorithms have been successfully applied to sentiment analysis and emotion recognition \cite{kastrati2024leveraging}.
	
	Elaborating on the characteristics of these models for our work, \textsc{nb} is a probabilistic classifier based on Bayes' theorem that estimates the probability of a given class based on a series of observed features. This model has proven effective in text classification tasks, where vector representations (such as bags of words or \textsc{tf}-\textsc{idf}) generate high-dimensional structures. Thanks to its low computational cost, ease of implementation, and competitive results, it can be used as a baseline. Moreover, \textsc{svm} is especially effective for high-dimensional data such as text, as it optimizes class separation through the superposition of hyperplanes. This enables robust classification even with small data sets. Its ability to handle multi-class problems makes it suitable for emotional analysis extended to a vast number of categories. \textsc{nn}s enable the capture of nonlinear and complex relationships in data. In text processing, this approach can learn hierarchical representations, detecting emotional nuances that elude simpler methods. Furthermore, \textsc{lr} is a linear classifier that is widely used for binary and multi-class classification. Its simplicity makes it a solid starting point, especially when combined with text representation techniques such as \textsc{tf}-\textsc{idf} or word embeddings. On the contrary, \textsc{knn} is a nonparametric method that classifies based on proximity in feature space. Although time-consuming to perform, it is helpful in exploratory phases and as a benchmark against more sophisticated models. Finally, boosting and gradient ensemble techniques combine multiple classifiers to build more accurate and robust models. They are instrumental when working with unbalanced distributions.
	
	Sentiment analysis enables automated and efficient processing of textual data to discern and categorize sentiment patterns. Typically, it focuses on determining the polarity of a text -- whether it expresses positive, negative, or neutral sentiment -- and is often used to measure public opinion, customer feedback, or overall sentiment towards a particular topic, product, or event \cite{nandwani2021review}. In this line, \cite{Zainuddin2014} applied \textsc{svm} to train a sentiment classifier with reference data sets. Moreover, \cite{Wehrmann2017} employed Convolutional Neural Networks (\textsc{cnn}s) for sentiment analysis. Unlike traditional language-dependent methods that rely on word-level, this language-agnostic model processes raw text at the character level. The resulting robust sentiment classifier works across multiple languages without extensive pre-processing or language-specific resources. Later, \cite{Wongkar2019} performed a sentiment analysis on Twitter data using the \textsc{nb} model. The probabilistic model categorized sentiment, providing valuable information on public opinion and trends on social media platforms. 
	
	Emotion recognition identifies the emotions contained within a text. 
	Unlike sentiment analysis, which typically categorizes text into broad sentiment categories (positive, negative, and neutral), emotion recognition aims to identify and categorize emotions such as sadness, joy, love, and anger. This more sophisticated analysis involves complex modeling to detect emotional cues \cite{nandwani2021review}. Accordingly, \cite{Parvin2021} used an \textsc{ml}-based ensemble technique to classify six primary textual emotions. Specifically, it compares eight standard ensemble techniques to conclude that the ensemble with Term Frequency-Inverse Document Frequency (\textsc{tf-idf}) achieves the best results. Moreover, \cite{liu2023emotion} proposed a Multi-label \textsc{knn} classifier to enable iterative adjustments in multi-label emotion recognition. This method was applied to enhance the accuracy and efficiency of emotion recognition in short Twitter texts. Among more recent works are the studies \cite{sujatha2025automatic} and \cite{tang2025speech}, which focused on emotion recognition from audio data. The study \cite{sujatha2025automatic} uses deep neural networks while \cite{tang2025speech} exploits \textsc{cnn}s.
	
	\subsection{Transformer models}
	
	\textsc{llm}s represent a significant evolution in deep learning applied to natural language. Based on transformer-like architectures, these models are capable of processing text sequences considering the full context of a sentence or paragraph, allowing for a richer and more accurate understanding of meaning. An \textsc{llm} is characterized by having been trained with large volumes of textual data, giving it a generalist capability to tackle multiple linguistic tasks. Typical applications include text generation, machine translation, sentiment analysis, and, more recently, emotion recognition \cite{Pico2024}.
	
	Linguistic feature extraction, used by initiatives such as the Semantic Orientation Calculator \cite{taboada2011lexicon}, assigns polarities to different words, creates a dictionary, and applies several algorithms to calculate emotional scores for each entry, resulting in a final classification. Recent advancements in \textsc{nlp} rely on developing transformer models and \textsc{llm}s \cite{ren2024advancements}. While a transformer model provides the underlying deep learning architecture, an \textsc{llm} applies the same architecture to complex linguistic tasks on a considerably larger scale. This is the case of well-known \textsc{llm}s like ChatGPT and Gemini, which have become the basis for many state-of-the-art \textsc{nlp} solutions. These models are dedicated to understanding and generating human language, including billions of parameters, and are trained with large amounts of text data in parallel.
	
	The mathematical operation of an \textsc{llm} begins with the conversion of the text into numeric tokens using a Byte Pair Encoding (\textsc{bpe}) segmentation scheme \cite{Li2025}. These tokens are transformed into fixed-dimensional vectors through an embedding layer, which assigns each token a continuous representation in a vector space. These vectors are processed sequentially by a stack of decoder blocks, each of which is responsible for progressively refining the internal representation of the sequence. The central component of each block is the Multi-Head Attention (\textsc{mha}) mechanism \cite{Kim2024}, which enables the model to determine the relevance of each token concerning the others within the sequence context. Mathematically, this mechanism projects the input vectors into three matrices: Query ($Q$), Key ($K$), and Value ($V$). The attention weights are then calculated using the scaled dot-product attention formula, which adjusts the magnitude of the similarities between $Q$ and transposed $K$ by dividing them by the square root of the dimension of $K$, thereby stabilizing the gradients and improving training efficiency. The final step is to apply the softmax function, which takes a vector of real values as input and converts them into a probability distribution.\\
	
	$\text{Attention}(Q, K, V) = \text{softmax}\left(\frac{QK^T}{\sqrt{d_k}}\right)V$\\
	
	The result of this operation is a weighted combination of the vectors $V$, in which the weights assigned to each position depend on its contextual relevance in the sequence. Now that each token has a contextual vector, it is passed to a Feed-Forward Network (\textsc{ffn}) model\footnote{Available at \url{https://arxiv.org/pdf/2406.08413}, September 2025.} \cite{Kim2024}, which is applied independently to each position in the sequence. This network is composed of two linear transformations separated by a nonlinear ReLU activation function ($\max(0,z)$). This component introduces nonlinearity and more complex transformation capability, allowing the model to represent highly expressive functions.\\
	
	$\text{FFN}(x) = \max(0, xW_1 + b_1)W_2 + b_2$\\
	
	Where $x$ is the input vector for a token, already enriched by the attention function; $W_1$ and $W_2$ are trainable weight matrices, whose values are adjusted during training using an optimization algorithm; and $b_1$ and $b_2$ are bias vectors that allow shifting the output and increasing the flexibility of the model.
	
	Currently, the most advanced and sophisticated solutions are based on transformers, which capture representations of words and contexts in a general and flexible way, adding significant value. Prominent models in emotion recognition include \textsc{bert} \cite{al2020emodet2} and \textsc{gpt} \cite{Mao2023}. Consideration should also be given to proprietary solutions, such as Anthropic, the basis for the enterprise conversational assistant Claude, or Inflection, a model used to create a personal intelligence assistant, as well as open source offers, such as Vicuna, a modified version of the \textsc{llm} by Meta, \textsc{ll}a\textsc{ma}.
	
	Transformer technology has been employed in various emotion recognition tasks, demonstrating superior performance in accurately identifying and classifying emotions across diverse data sets and languages \cite{Mao2023,peng2022survey,Wake2023}. This is true for both multimodal environments -- including multiple types of data like audio, video, and text \cite{Cheng2024,Ma2024,Peng2024} -- and unimodal environments -- which rely on a single data type \cite{Pico2024,Lei2024,Venkatesh2024,zhang2025dialoguellm,hong2025aer}. Consequently, transformer models can perform audio-based, video-based, and text-based emotion recognition.
	
	\cite{Cheng2024} proposed Emotion-\textsc{ll}a\textsc{ma}, a large multimodal language model. It incorporates Hu\textsc{bert} for audio processing and visual encoders to gather facial details, dynamics, and context. Emotion-\textsc{ll}a\textsc{ma} significantly enhances emotional recognition and reasoning capabilities by integrating multiple descriptive elements like the audio tone, lexical subtitle, visual objective, visual expression, classification label, and modality. Conversely, \cite{Peng2024} applied model adaptation techniques -- deep prompt tuning and low-rank adaptation -- to customize the Chat General Language Model (Chat\textsc{glm}), an open-source pre-trained language model, for emotion recognition tasks. The adapted versions outperform state-of-the-art models, tested on six audio, video, and text datasets.
	
	More recent solutions using transformers are \cite{hussain2025low,yi2025hyfuser,fu2025laerc}. Firstly, \cite{hussain2025low} proposed MobileBERT for emotion recognition from textual and video data. A similar solution is proposed in \cite{yi2025hyfuser}, which utilizes KoELECTRA and HuBERT in a multimodal scenario involving both textual and audio data. Finally, in \cite{fu2025laerc}, a textual emotion recognition system based on LLaMA2 is presented.
	
	Moreover, Emotion Recognition in Conversation (\textsc{erc}) focuses on detecting emotions during dialogues. It aims to identify the emotional category of each utterance in a conversation, whether text-based or audio-based. In this line, \cite{Lei2024} introduced Instruct\textsc{erc}, a framework that combines the strengths of retrieval-augmented mechanisms and \textsc{llm} solutions like \textsc{gpt}-3 and \textsc{t}5 to access external knowledge and contextual information, thereby addressing the limitations of traditional \textsc{erc} models. By employing these models, Instruct\textsc{erc} improves the accuracy of emotion classification in dialogues. Conversely, \cite{Pico2024} explored the text-generating capabilities of \textsc{llm}s to enrich intelligent conversational agents with the ability to recognize and adapt to the emotions of the partner speaker during textual dialogues. Similarly, \cite{Venkatesh2024} proposed a text-based \textsc{erc} considering contents and contextual factors like dialogue history, speaker roles, and the interplay between different conversational turns. In the end, \cite{zhang2025dialoguellm} presented Dialogue\textsc{llm}, an emotion and context knowledge enhanced language model designed explicitly for \textsc{erc}, based on open-source base models, namely \textsc{ll}a\textsc{ma}2. Similarly, \cite{hong2025aer} and \cite{bo2025toward} presented new advancements in the field of \textsc{erc}. More in detail, \cite{hong2025aer} worked with ambiguous emotions in zero-shot and few-shot settings, while \cite{bo2025toward} focused on zero-shot conditions but performed experiments with real and synthetic data in both text and speech modalities.
	
	Speech Emotion Recognition (\textsc{ser}) identifies and classifies the speaker's emotional state based on vocal expressions. This implies analyzing various speech signal features, such as pitch, tone, intensity, rhythm, and prosody, to detect emotions like happiness, sadness, anger, and fear. In this line, \cite{Ma2024} explored the integration of speech analysis, text generation, and speech synthesis. The data2vec pre-trained model performs speech analysis to capture nuanced vocal features; \textsc{gpt}-4 generates text to provide contextual understanding and augment emotion detection, and Azure Text-to-Speech implements emotional speech synthesis to create a more expressive and accurate \textsc{ser} system.
	
	Finally, recognizing emotions through transformer technology enables asking open-ended questions using an old human strategy, \textit{i.e.}. An open-ended query expands the range of potential responses and increases the model's uncertainty. The model generates an elongated response to traverse the spectrum of possible interpretations to curtail this ambiguity. In this regard, \cite{Amirizaniani2024} evaluated the capabilities of \textsc{llm}s to understand human intentions, emotions, and reasoning processes when addressing open-ended questions. The study compares human and \textsc{llm} responses using Zephyr-7B, \textsc{ll}a\textsc{ma}2, and \textsc{gpt}-4. The results show the effectiveness of incorporating mental states, such as human intentions and emotions, into prompt tuning to improve the quality of \textsc{llm} reasoning. Moreover, our prior work by \cite{Pajon2024} combined contextual information with prompt engineering and a general-purpose \textsc{llm} to enhance emotion recognition. We adopted a prompt template integrating the head, emotions, polarities, objective, structure, question, and optional conversation. In addition, \cite{Peng2024} recommended and implemented a two-sentence prompt template. The first sentence provides the emotion recognition instructions: ``\texttt{\small Classify the sentiment of the sentence to Emotion\textsubscript{1}, Emotion\textsubscript{2}, ..., Emotion$_{k}$}''. The second sentence holds the contents submitted for emotion recognition: ``\texttt{\small <a single sentence from the test set>}''. The value of $k$ corresponds to the number of sentiment/emotion categories specific to the dataset.
	
	In summary, most of the reviewed research follows code rather than prompt-oriented strategies, which significantly restricts reuse by other researchers and professionals. Standard code-oriented strategies can be used with specialized and general-purpose pre-trained models, whereas prompt-oriented strategies only work with general-purpose pre-trained models. The primary advantages of the current prompt-based proposal over existing methods lie in its simplicity and seamless adaptability to diverse fields and general-purpose pre-trained models. 
	
	\subsection{Research contributions}
	\label{sec:research_contribution}
	
	Emotion recognition has improved human-computer interaction, enhanced customer service, and supported mental health interventions. It involves identifying and classifying emotional states from textual, auditory, or visual data, making it essential for creating empathetic and responsive intelligent systems.
	This work presents a novel contribution to the field of emotion recognition from open text through the combined use of specialized pre-trained models and general-purpose \textsc{llm}s. Unlike previous studies that focus exclusively on a single technique (either fine-tuning or prompt engineering), our proposal systematically compares both approaches in different scenarios, also incorporating different prompt design and emotion grouping strategies. The main contributions are the following:
	
	\begin{figure*}[!htbp]
		\centering
		\includegraphics[width=0.6\textwidth]{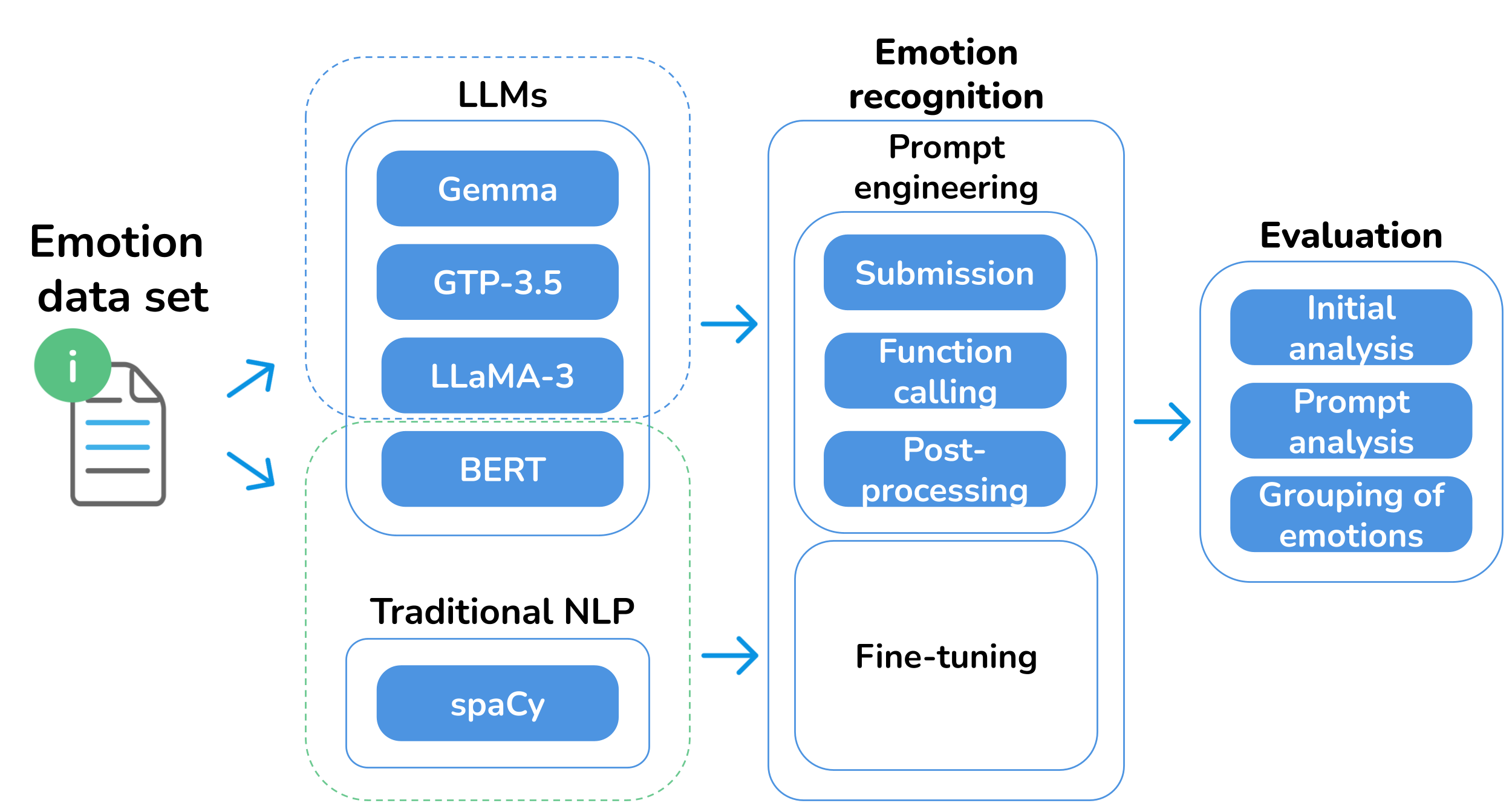}
		\caption{System diagram.}
		\label{fig:system_diagram}
	\end{figure*}
	
	\begin{itemize}
		\item A hybrid approach is proposed that integrates fine-tuned models and general models with prompt engineering techniques.
		
		\item Five types of prompts and three different ways of grouping emotions are analyzed, evaluating their impact on model performance.
		
		\item The approach's replicability and extension are facilitated by providing reusable prompt structures that are adaptable to different models.
	\end{itemize}
	
	Table \ref{tab:comparison_literature} offers a comparative analysis of the emotion recognition works referred to above considering the task (emotion classification: \textsc{ec}, emotion recognition: \textsc{er}, \textsc{erc}, sentiment analysis: \textsc{sa} and \textsc{ser}), the data modality (audio: \textsc{a}, text: \textsc{t}, video: \textsc{v}), the category of the technique (traditional \textsc{ml} or transformer-based) and the usage of prompt-based strategies. 
	
	\begin{table*}[!htbp]
		\centering
		\footnotesize
		\caption{Comparison of the surveyed works concerning emotion analysis with ML and transformer models.}
		\label{tab:comparison_literature}
		\begin{tabular}{lcccccc}
			\toprule
			\multirow{2}{*}{\textbf{Authorship}} & \multirow{2}{*}{\textbf{Task}} & \textbf{Data} & \multirow{2}{*}{\textbf{Technique}} & \textbf{Prompt} \\
			& & \textbf{modality} & & \textbf{engineering} \\
			\midrule
			\cite{Zainuddin2014} & \multirow{3}{*}{\textsc{sa}} & \multirow{3}{*}{\textsc{t}} & \multirow{3}{*}{\textsc{ml}} & \multirow{3}{*}{\xmark}\\
			\cite{Wehrmann2017}\\ 
			\cite{Wongkar2019}\\
			\midrule
			\cite{Ma2024} & \textsc{ser} & \textsc{a,v} & Transformer & \xmark\\
			\midrule
			\cite{Parvin2021} & \multirow{2}{*}{\textsc{ec}} & \multirow{2}{*}{\textsc{t}} & \multirow{2}{*}{\textsc{ml}} & \multirow{2}{*}{\xmark} \\
			\cite{liu2023emotion}\\
			\midrule
			
			\cite{sujatha2025automatic} & \multirow{2}{*}{\textsc{er}} & \multirow{2}{*}{\textsc{a}} & \multirow{2}{*}{\textsc{ml}} & \multirow{2}{*}{\xmark} \\
			\cite{tang2025speech}\\
			\midrule
			
			\cite{Mao2023} & \textsc{er}, \textsc{sa} & \textsc{t} & \multirow{11}{*}{Transformer} & \xmark\\
			
			\cite{Wake2023}& \multirow{10}{*}{\textsc{er}} & \textsc{t} & & \xmark \\
			
			\cite{hussain2025low} & & \textsc{t,v} & & \xmark \\
			\cite{yi2025hyfuser} & & \textsc{a,t} & & \xmark \\
			\cite{fu2025laerc} & & \textsc{t} & & \cmark \\
			
			\cite{Amirizaniani2024} & & \textsc{t} & & \cmark \\
			\cite{Cheng2024} & & \textsc{a,v} & & \xmark \\
			\cite{Pajon2024} & & \textsc{t} & & \cmark \\
			\cite{Peng2024} & & \textsc{a,t,v} & & \cmark \\
			\cite{Pico2024} & & \textsc{t} & & \xmark \\
			\cite{Venkatesh2024} & & \textsc{t} & & \xmark \\
			\midrule
			
			\cite{Lei2024} & \multirow{4}{*}{\textsc{erc}} & \textsc{t} & \multirow{4}{*}{Transformer} & \xmark\\
			\cite{zhang2025dialoguellm} & & \textsc{t} & & \xmark\\
			\cite{hong2025aer} & & \textsc{t} & & \cmark\\
			\cite{bo2025toward} & & \textsc{a,t} & & \cmark\\
			\midrule
			
			\bf \multirow{2}{*}{Proposed solution} & \multirow{2}{*}{\textsc{er}} & \multirow{2}{*}{\textsc{t}} & \textsc{ml} & \multirow{2}{*}{\cmark} \\
			& & & Transformer & \\
			
			\bottomrule
		\end{tabular}
	\end{table*}
	
	While earlier works, such as those by \cite{Zainuddin2014}, \cite{Wehrmann2017}, and \cite{Wongkar2019}, focused on sentiment analysis using \textsc{ml}-based techniques, more recent studies adopt transformer-based methods for text-based \cite{Pico2024,Mao2023,Wake2023,Lei2024,Venkatesh2024,Amirizaniani2024,Pajon2024,zhang2025dialoguellm,hong2025aer} and multimodal \cite{Cheng2024,Ma2024,Peng2024,bo2025toward} emotion detection. Only  \cite{Peng2024}, \cite{Amirizaniani2024}, and  \cite{Pajon2024} have explored prompt-based strategies for emotion recognition with transformer models. \cite{Amirizaniani2024} used prompts to perform text-based emotion recognition. Specifically, they compare, for a given topic, the reasoning quality and the emotional contents of open-ended responses produced by humans and \textsc{llm} models. \cite{Pajon2024} used prompt engineering with \textsc{gpt}-3.5 for contextual emotion recognition within interactive conversations and extensive texts. The extensive texts are dynamically divided into fragments and submitted sequentially as a conversation. In the case of a conversation, the prompt holds the entire conversation as context and indicates which part to analyze in each iteration; in the case of an extensive text, the prompt holds the emotions and polarities identified so far as context and specifies the text fragment to analyze in each iteration. This decision to provide the maximum possible context enables the model to achieve more profound results, as past events are essential. \cite{Peng2024} adapted the pre-trained Chat\textsc{glm} language model for emotion recognition and then tested the resulting models using a sentence-level emotion recognition prompt. 
	
	In contrast, the proposed solution integrates \textsc{ml}- and transformer-based techniques with multiple prompt strategies and emotion groupings. The design and refinement of open-ended questions are innovative features that enhance the model's ability to adapt and respond to contextual variations, significantly advancing emotion recognition. Furthermore, the current work leverages the power of fine-tuned specialized and general-purpose pre-trained models to enhance the performance of text-based emotion recognition.
	
	\section{Method}
	\label{sec:proposed_method}
	
	Figure \ref{fig:system_diagram} presents the modules of the proposed architecture. This work distinguishes between traditional and \textsc{llm} approaches, dividing \textsc{llm} into general- and specific-purpose models. To this end, a general-purpose \textsc{llm} provides an already pre-trained model to evaluate the performance of prompt engineering.
	
	\subsection{Large Language Models}
	\label{sec:llms}
	
	This work explores the following \textsc{llm} implementations: Gemma, \textsc{gpt}-3.5, \textsc{ll}a\textsc{ma}-3, and \textsc{bert}.
	
	\subsubsection{Gemma}
	\label{sec:gemma}
	
	Gemma is a model developed by Google and introduced in February 2024. The adopted Gemma 1.1 model (gemma1.1-7b-it\footnote{Available at \url{https://huggingface.co/docs/transformers/model_doc/gemma}, September 2025.}) has 7 billion parameters and offers excellent versatility in a wide range of areas. This version (1.1) has undergone substantial modifications by being trained using a novel method of Reinforcement Learning from Human Feedback (\textsc{rlhf}). This resulted in significant improvements in quality, coding capabilities, veracity, instruction following, and quality of multi-turn conversations. 
	
	The great advantage of this model lies in the balance between computing capacity and the resources required. Although it has a limited number of parameters compared to others, it exhibits very high performance in various applications.
	
	\subsubsection{GPT-3.5}
	\label{sec:gpt}
	
	\textsc{gpt}-3.5\footnote{Available at \url{https://platform.openai.com/docs/models/gpt-3-5-turbo}, September 2025.}, based on \textsc{gpt}-3, represents a significant evolution in natural language generation technology. This model, created by OpenAI, demonstrates an enhanced ability to comprehend and generate text with a deeper and more coherent context, thanks to its use of 175 billion parameters and extensive training on diverse datasets.
	
	Compared to other models, the distinctive features of \textsc{gpt}-3.5 are its scale and complexity, which translate into high fluency and a deep understanding of complex linguistic features. This last feature allows this version to explore broader contexts than its predecessors. 
	
	Thanks to these improvements, the \textsc{gpt}-3.5 model can be utilized in complex domains and real-time applications due to its enhanced inference capability, which enables faster responses. This makes it ideal for complex process automation tasks, such as the current work.
	
	\subsubsection{LLaMA-3}
	\label{sec:llama}
	
	\textsc{ll}a\textsc{ma}-3 is a model developed by Meta and introduced in April 2024. The third and most recent version includes powerful \textsc{nlp} capabilities similar to those of the models mentioned above.
	
	The current system uses the 8 billion parameters version oriented to instructions\footnote{Available at \url{https://huggingface.co/meta-llama/Meta-Llama-3-8B-Instruct}, September 2025.} to provide optimized outputs for feature extraction processes. Unlike other models, 
	\textsc{ll}a\textsc{ma}-3 needs to incorporate different prompts delimited by keywords indicated in the documentation to exemplify the outputs.
	
	\subsubsection{BERT}
	\label{sec:bert}
	
	\textsc{bert} is a language model developed by Google in 2018. Unlike traditional \textsc{nlp} models, which process text sequentially and in only one direction (left-to-right or right-to-left), \textsc{bert} employs a transformer architecture that enables bidirectional encoding of context. This means that \textsc{bert} can simultaneously consider information from both preceding and following words in a sentence, providing a richer and more accurate understanding of the contextual meaning of each word.
	
	One of the most prominent features of \textsc{bert} is its ability to pre-train large corpora of unlabelled text using two main tasks: masked language modeling (\textsc{mlm}) and next sentence prediction (\textsc{nsp}). With \textsc{mlm}, \textsc{bert} learns to predict hidden words in a sentence based on context, while with \textsc{nsp}, the model learns the relationship between two consecutive sentences.
	
	The versatility of \textsc{bert} lies in its ability to adjust to specific \textsc{nlp} tasks with little additional labeled data. This feature allowed us to fine-tune \textsc{bert} for emotion recognition. The process ensures that the system consistently produces an adequate output, regardless of the quality of a specific prompt.
	
	\subsubsection{RoBERTa}
	\label{sec:roberta}
	
	\textsc{r}o\textsc{bert}a is an optimized variant of \textsc{bert}, introduced by Facebook \textsc{ai} in 2019. 
	Unlike the original \textsc{bert}, \textsc{r}o\textsc{bert}a removes the next-sentence prediction objective, 
	applies dynamic masking during pretraining, and is trained on a significantly larger corpus. 
	These modifications lead to richer contextual representations and better generalization capabilities, 
	which are crucial for fine-grained emotion recognition. Consequently, \textsc{r}o\textsc{bert}a has been widely adopted as one of the reference transformer models in multiple \textsc{nlp} tasks, including sentiment and emotion analysis \cite{alqarni2025emotion}. To avoid redundancy in diagrams and figures, only \textsc{bert} is explicitly depicted since \textsc{r}o\textsc{bert}a is a direct improvement, sharing the same underlying architecture with modifications in the pretraining process (e.g., dynamic masking and removal of the next-sentence prediction objective). Therefore, whenever \textsc{bert} is referenced in methodological diagrams, it should be understood that \textsc{r}o\textsc{bert}a is also encompassed in the evaluation.
	
	\subsection{Traditional models}
	\label{sec:traditional_approach}
	
	spaCy is an \textsc{nlp} library written in Python, which stands out for speed, efficiency, and accuracy in a wide range of \textsc{nlp} tasks. Spacy allows the training of small models focused on a specific classification function, like detecting six emotion classes\footnote{Available at \url{https://spacy.io/api/textcategorizer}, September 2025.}.
	
	Spacy is a powerful tool for building a text classification model for polarity or entity detection. These features, combined with the extensive existing community, make SpaCy a versatile tool for text-based emotion detection.
	
	\subsection{Emotion recognition}
	\label{sec:emotional_analysis}
	
	\subsubsection{Prompt engineering}
	\label{sec:prompt}
	
	The current prompt engineering approach is applied to Gemma, \textsc{gpt}-3.5, and \textsc{ll}a\textsc{ma}-3. The prompts used with Gemma (see Listing \ref{gemma_prompt} for an illustrative example) and \textsc{gpt}-3.5 (see Listing \ref{gpt_prompt}) are similar because the system understands the instructions perfectly without needing special commands. The \textsc{ll}a\textsc{ma}-3 prompt requires, as referred to in Section \ref{sec:llama}, specific keywords to obtain optimal results (see Listing \ref{llama_prompt}). These prompts provide context (\texttt{\small Contxt}), instructions (\texttt{\small Instru}), and the sentence (\texttt{\small Sentnc}) for the general-purpose \textsc{llm} to analyze.\\
	
	\begin{figure*}[htbp]
		\centering
		\begin{minipage}{1.0\textwidth}
			\captionof{lstlisting}{Gemma prompt and answer template.}\label{gemma_prompt}
			\begin{lstlisting}[]
				*Contxt:* Imagine that you are doing a study about the emotions present in a text. 
				*Instru:* Only detect the following emotions in the study: sadness, joy, love, anger, fear, surprise. 
				*Instru:* Objective: Detect the key emotion present in the text. 
				*Instru:* The output will be a JSON list with the key emotion delimited by % like %{value:x}%. 
				*Instru:* Perform the study of this text fragment following strictly the structure indicated above, without introducing any of the given text and only with the emotions indicated -> 
				*Sentnc:* sentence
				*Answer:* %{value:x}%
			\end{lstlisting}
		\end{minipage}
	\end{figure*}
	
	\begin{figure*}[htbp]
		\centering
		\begin{minipage}{1.0\textwidth}
			\captionof{lstlisting}{GPT-3.5 prompt and answer template.}\label{gpt_prompt}
			\begin{lstlisting}[]
				*Contxt:* You are doing an emotional study on text input.
				*Instru:* You will organize the emotions in 3 independent groups focusing on the emotion you want to express: the positive emotion group will be (love), the negative emotion group will be (fear), and the neutral emotion group will be (surprise). 
				*Instru:* The output will be a JSON list with a single key with the format -> emotion: positive, negative, or neutral. 
				*Instru:* The input text will be enclosed in three quotes.
				*Sentnc:* '''sentence'''
				*Answer:* emotion: value
			\end{lstlisting}
		\end{minipage}
	\end{figure*}
	
	\begin{figure*}[htbp]
		\centering
		\begin{minipage}{1.0\textwidth}
			\captionof{lstlisting}{\textsc{ll}a\textsc{ma}-3 prompt and answer template.}\label{llama_prompt}
			\begin{lstlisting}[]
				*Instru:* <|begin_of_text|><|start_header_id|>system<|end_header_id|>You are a system that always detects the emotions [sadness, joy, love, anger, fear, surprise]. The answer format will only include the detected emotion. Never mix two emotions; only make single detections. 
				*Contxt:* All texts are for academic study.<|eot_id|>
				*Sentnc:* <|start_header_id|> user <|end_header_id|>sentence<|eot_id|>
				*Answer:* <|start_header_id|>assistant<|end_header_id|>value<|eot_id|> 
			\end{lstlisting}
		\end{minipage}
	\end{figure*}
	
	Prompt execution comprises prompt submission, with the support of function calling in the case of \textsc{gpt}-3.5, and response post-processing. The goal is to identify the specified emotions within the submitted text. 
	\begin{itemize}
		\item Submission. The interface with Gemma was provided via a Python server with Flask\footnote{Available at \url{https://flask.palletsprojects.com/en/3.0.x}, September 2025.} as front-end and the gemma1.1-7b-it\footnote{Available at \url{https://huggingface.co/google/gemma-1.1-7b-it}, September 2025.} model as a back-end. The \textsc{ll}a\textsc{ma}-3 model was deployed using the Text Generation Interface\footnote{Available at \url{https://huggingface.co/docs/text-generation-inference/index}, September 2025.} system provided by Hugginface to optimize model generation runtime. However, it is not yet fully compatible with Gemma. 
		Both models have been deployed using Kubernetes, allowing a fast and resource-efficient deployment.
		The interface with \textsc{gpt}-3.5 was via the \textsc{api} offered by OpenAI. The calls to the \textsc{api} require using an \textsc{api}-key. Being \textsc{gpt}-3.5, a private model, OpenAI implements a pay-as-you-go charging model.
		
		\item Function calling is used with \textsc{gpt}-3.5. This functionality allows the specification of the expected outputs. This provides greater precision when making requests, avoiding results with additional text or wrong ones. However, it is still necessary to incorporate post-processing functions to obtain a clean output according to the desired format. 
		
		\item Post-processing is applied to \textsc{llm} outputs to filter out unwanted words and incorrect formatting. In some cases, emotions with similar connotations can be confused, \textit{e.g.}, joy and hope. Therefore, one of the post-processing steps is to normalize the values using a dictionary regarding the six considered emotions\footnote{Available at \url{https://bit.ly/3W2TFpB}, September 2025.}.
		Using model-specific regular expressions\footnote{Available at \url{https://bit.ly/3XYMNfD}, September 2025.} helps to remove unwanted characters and words, \textit{e.g.}, line breaks or quotation marks, or the word \textsc{json} that appears explicitly with Gemma. 
		
	\end{itemize}
	
	\subsubsection{Fine-tuning}
	\label{sec:bert_module}
	
	As fine-tuning of Gemma, \textsc{gpt}-3.5 and \textsc{ll}a\textsc{ma}-3 is out of the scope of this work, this step was applied just to spaCy and \textsc{bert}.
	
	Being a transformer model, \textsc{bert} was fine-tuned to recognize the six emotions. First, it was trained using the aforementioned fine-tuning data partition with the parameters in Table \ref{bert-params}. Table \ref{roberta-params} details the parameters used in the \textsc{r}o\textsc{bert}a model. This training produces a reduced model that can be executed on machines with limited resources. 
	
	spaCy is a traditional model that incorporates a series of functionalities, allowing for adaptation to any field or application. This is the case of pipes, which support model retraining for emotion recognition. Specifically, this work uses a test categorizer of six classes corresponding to the target emotions. Subsequently, pretraining was performed with the fine-tuning data partition described in Section \ref{subsec:dataset} to detect the six target classes solely. 
	
	\begin{table}[!htbp]
		\centering
		\caption{BERT parameters.}
		\begin{tabular}{lr}
			\hline
			\textbf{Parameter} & \textbf{Value} \\ \hline
			num\_classes & 6 \\ 
			max\_length & 128 \\ 
			batch\_size & 16 \\ 
			num\_epochs & 4 \\ 
			learning\_rate & 2e-5 \\ \hline
		\end{tabular}
		\label{bert-params}
	\end{table}
	
	\begin{table}[!htbp]
		\centering
		\caption{RoBERTa parameters.}
		\begin{tabular}{lr}
			\hline
			\textbf{Parameter} & \textbf{Value} \\ \hline
			num\_classes & 6 \\ 
			max\_length & 128 \\ 
			batch\_size & 32 \\ 
			num\_epochs & 8 \\ 
			learning\_rate & 1e-5 \\ \hline
		\end{tabular}
		\label{roberta-params}
	\end{table}
	
	Finally, the fine-tuned \textsc{bert}, \textsc{r}o\textsc{bert}a and spaCy models are ready for evaluation using the fine-tuning data partition also described in Section \ref{subsec:dataset}. Note that neither \textsc{bert}, \textsc{r}o\textsc{bert}a nor spaCy supports user prompting.
	
	\subsubsection{Evaluation}
	\label{sec:evaluation}
	
	The results are evaluated using classical Artificial Intelligence metrics, such as accuracy, recall, precision, and \textsc{f}-score. 
	
	To analyze the behavior of the selected models with different prompts and emotion groups, there are experiments with five types of prompts (see Section \ref{sec:scenarios}) and three emotion groupings with six, three, and two classes (see Table \ref{tab:groupings}). The six classes correspond to the sadness, joy, love, anger, fear, and surprise emotions; the three classes refer to positive (love), negative (fear), and neutral (surprise), whereas the two classes correspond to positive (joy/love) and negative (anger/sadness) feelings. The class grouping was based on the emotional charge. 
	
	\begin{table*}[!htbp]
		\centering
		\caption{Emotion groupings.}
		\begin{tabular}{cll}
			\toprule
			(\#) & Classes & Emotions \\ \midrule
			6 & sadness, joy, love, anger, fear, surprise & sadness, joy, love, anger, fear, surprise\\ 
			3 & positive, negative, neutral & love, fear, surprise\\ 
			2 & positive, negative & joy/love, anger/sadness\\ \bottomrule
		\end{tabular}
		\label{tab:groupings}
	\end{table*}
	
	\subsection{Scenarios}
	\label{sec:scenarios}
	
	The experiments consider three scenarios:
	\begin{itemize}
		\item S1 compares the performance of general-purpose \textsc{llm}s versus pre-trained models using basic prompts (see Section \ref{sec:prompt}).
		
		\item S2 analyses the performance of the \textsc{llm}s with different prompts: 
		\begin{itemize}
			\item Basic prompt requests a single emotion from the available lists (see Section \ref{sec:prompt}).
			\item Mask prompt applies a binary mask to detect the emotions (see Table \ref{tab:prompt_mask}).
			\item Percent prompt requests emotion percentages in \textsc{json} format. The model will be instructed that the desired output is a \textsc{json} list with the percentage of each analyzed emotion in the text. It will also be established that there must always be a dominant emotion.
			\item Numerical prompt associates each emotion with a number.
			\item Inverse prompt asks for the inverse emotion. The model is instructed to identify the inverse emotion to the one in the text. This prompt establishes whether the model can make complex associations between emotions.
		\end{itemize}
		\item S3 assesses the impact of the distribution, choice, and grouping of emotions (six, three, and two classes) in the \textsc{llm}s.
		
	\end{itemize}
	
	\begin{table}[!htbp]
		\centering
		\caption{Relation between emotions and masks.}
		\begin{tabular}{ll}
			\hline
			Emotion & Mask \\ \hline
			sadness & 000001 \\ 
			joy & 000010 \\ 
			love & 000100 \\ 
			anger & 001000 \\ 
			fear & 010000 \\ 
			surprise & 100000 \\ \hline
		\end{tabular}
		\label{tab:prompt_mask}
	\end{table}
	
	\section{Experimental results}
	\label{sec:experimental_results}
	
	Experiments were performed on a computer with the following hardware specifications:
	
	\begin{itemize}
		\item [--] \textbf{Operating System}: Ubuntu 22.04.4 LTS 64 bits.
		\item [--] \textbf{Processor}: Intel\@Core i7-13700K \SI{3.40}{\giga\hertz}.
		\item [--] \textbf{RAM}: \SI{32}{\giga\byte} DDR4.
		\item [--] \textbf{Disk}: \SI{1000}{\giga\byte} NVME.
		\item [--] \textbf{GPU}: Nvidia GTX-1050Ti \SI{4}{\giga\byte}.
	\end{itemize}
	
	\noindent The \textsc{llm} experiments were performed in a server with the following hardware specifications:
	
	\begin{itemize}
		\item [--] \textbf{Operating System}: Debian 10 Buster 64 bits.
		\item [--] \textbf{Processor}: Intel\@Xeon Gold 5317 \SI{3.00}{\giga\hertz}.
		\item [--] \textbf{RAM}: \SI{128}{\giga\byte} DDR4.
		\item [--] \textbf{Disk}: \SI{100}{\giga\byte} SSD.
		\item [--] \textbf{GPU}: Nvidia A10 \SI{20}{\giga\byte}.
	\end{itemize}
	
	\subsection{Experimental data sets}
	\label{subsec:dataset}
	
	The experimental data is publicly available\footnote{Available at \url{https://www.kaggle.com/datasets/parulpandey/emotion-dataset}, September 2025.}. Table \ref{tab:dataset_distribution} shows the distribution by emotion category of two subsets (train and test). The first is used to fine-tune the traditional and \textsc{bert} models. The second, with \num{16000} samples, is intended for their evaluation.
	
	\begin{table}[!htbp]
		\centering
		\caption{\label{tab:dataset_distribution}Experimental data set.}
		\begin{tabular}{llS[table-format=6.0]}
			\toprule 
			\bf Partition & \textbf{Class} & \multicolumn{1}{c}{\textbf{Number of entries}}\\ \midrule
			\multirow{7}{*}{Fine-tuning} 
			& sadness & 581\\
			& joy & 695\\
			& love & 159\\
			& anger & 275\\
			& fear & 224\\
			& surprise & 66\\
			\cmidrule{2-3}
			& \bf Total & 2000\\
			\midrule
			\multirow{7}{*}{Evaluation} 
			& sadness & 4666\\
			& joy & 5362\\
			& love & 1304\\
			& anger & 2159\\
			& fear & 1937\\
			& surprise & 572\\
			\cmidrule{2-3}
			& \bf Total & 16000\\
			\bottomrule
		\end{tabular}
	\end{table}
	
	\subsection{Theoretical evaluation}
	
	In this section, a theoretical evaluation of the scenarios S1, S2, and S3 presented in Section \ref{sec:scenarios} is performed.
	
	\subsubsection{S1: \textsc{llm}s \textit{versus} pre-trained models}
	
	Being \( m \) a model (either an \textsc{llm} or a pre-trained model or \textsc{pre}) and \( p_b \) a basic prompt, an evaluation metric (\textit{e.g.}, accuracy, \textsc{f}-score) is defined as: 
	
	\[
	\mathcal{M}(m, p_b)
	\]
	
	The performance difference between \textsc{llm}s and pre-trained models can be expressed as:
	
	\[
	\Delta_\mathcal{M} = E_{m \in \\LLM} \left[ \mathcal{M}(m, p_b) \right] - E_{m' \in \\PRE} \left[ \mathcal{M}(m', p_b) \right]
	\]
	
	where
	\[
	E_{m \in \mathcal{A}} \left[ \mathcal{M}(m, p_b) \right] = 
	\frac{1}{|\mathcal{A}|} \sum_{m \in \mathcal{A}} \mathcal{M}(m, p_b)
	\]
	
	\subsubsection{S2: prompt comparison}
	
	Given a model \( m \), different prompt strategies are evaluated \( p \in \mathcal{P} = \{p_b, p_m, p_\%, p_n, p_\text{inv} \} \), where:
	
	\begin{itemize}
		\item \( p_b \) is the basic prompt.
		\item \( p_m \) is the binary-masked prompt.
		\item \( p_\% \) is the prompt based on percentages in \textsc{json} format.
		\item \( p_n \) is the numerical codified prompt.
		\item\( p_\text{inv} \) is the inverse emotion prompt.
	\end{itemize}
	
	The difference in performance between the two prompt strategies is defined as:
	
	\[
	\Delta_{\mathcal{M}i,j} = \mathcal{M}(m, p_i) - \mathcal{M}(m, p_j) \quad \forall p_i, p_j \in \mathcal{P},\, i \ne j
	\]
	
	\subsubsection{S3: emotion grouping evaluation}
	
	Being \( \mathcal{C} \) the set of emotional classess and \( \Pi_k \) a partition of \( \mathcal{C} \) in \( k \) non-empty subsets (in our study \( k \) = \{6,3,2\}), \( \Pi_k \) fulfills that
	
	\[
	\forall A \in \Pi_{k}, \, \exists B \in \Pi_{k'} / A \subseteq B, where
	\] \[ k > k'
	\]
	
	This relationship follows the partition refinement principle, where \( \Pi_{k} \) is a refinement of \( \Pi_{k'} \), which implies a higher emotional granularity.
	
	For a model \( m \) and a prompt \( p \), we define the performance as \( \mathcal{M}(m, p, \Pi_k) \). The difference in performance from a higher to a lower granularity, that is, grouping emotions, is defined as:
	
	\[
	\Delta_{\mathcal{M}k, k'} = \mathcal{M}(m, p, \Pi_{k'}) - \mathcal{M}(m, p, \Pi_{k})
	\]
	
	This value represents the gain obtained by reducing the class space complexity. Under the specific emotion separability hypothesis, we may assume that if \( H(\Pi_k) \) is the entropy associated with the portion \( \Pi_k \).
	
	\[
	H(\Pi_{k})>H(\Pi_{k'})
	\]
	
	This formulation enables us to assess the impact on output space reduction as a supervised semantic compression phenomenon using set theory and information theory.
	
	\subsection{Experimental evaluation}
	
\begin{figure*}[!htbp]
		\centering
		
		\subfloat[\label{fig:full_gemma} \centering Gemma]
		{{\includegraphics[width=6cm]{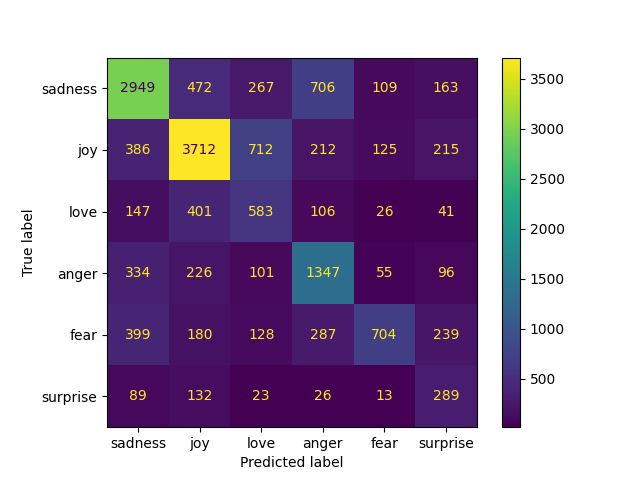}}} 
		\,
		\subfloat[\label{fig:full_gpt} \centering \textsc{gpt}-3.5]
		{{\includegraphics[width=6cm]{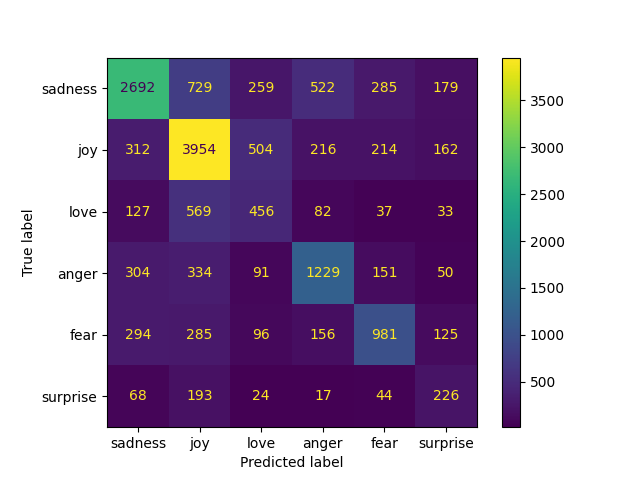}}} 
		\,
		\subfloat[\label{fig:full_llama} \centering \textsc{ll}a\textsc{ma}-3]
		{{\includegraphics[width=6cm]{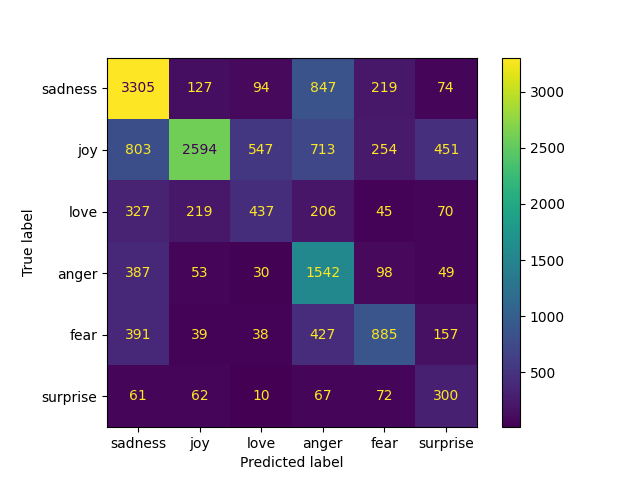}}}
		\,
		\subfloat[\label{fig:full_bert} \centering \textsc{bert}]
		{{\includegraphics[width=6cm]{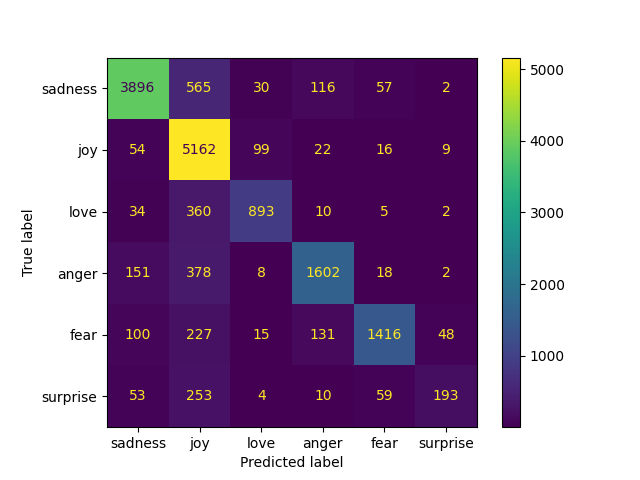}}}
		\,
		\subfloat[\label{fig:full_spacy} \centering spaCy]
		{{\includegraphics[width=6cm]{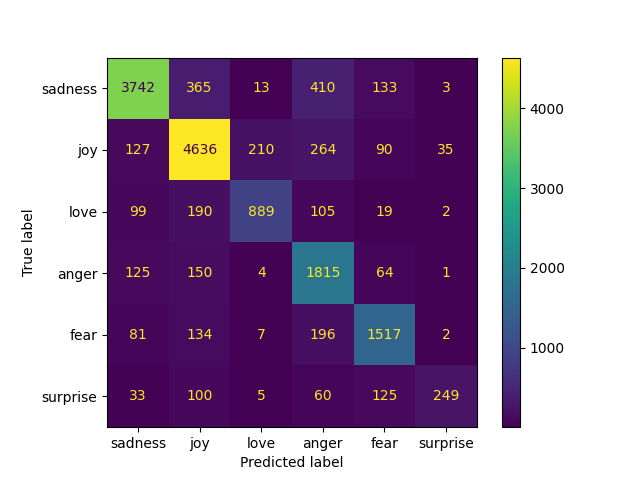}}}
		\,
		\subfloat[\label{fig:full_roberta} \centering \textsc{r}o\textsc{bert}a]
		{{\includegraphics[width=6cm]{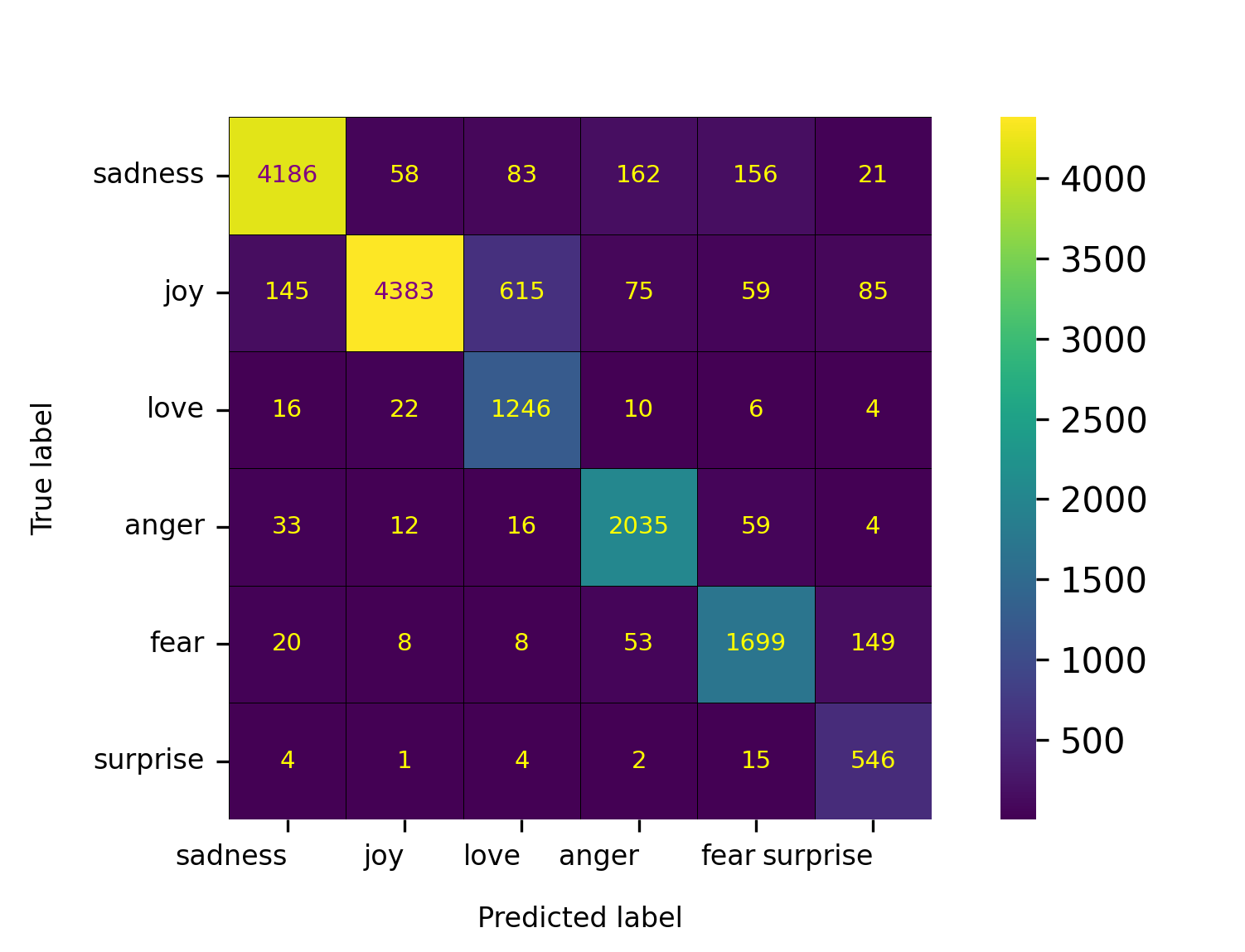}}}
		\caption{Comparison between fine-tuned and general-purpose models.}
		\label{fig:retrain_vs_llms}
	\end{figure*}

	Table \ref{tab:comparison} shows the results of the first scenario. The fine-tuning approaches (\textsc{bert}, spaCy, and especially \textsc{r}o\textsc{bert}a) yield the best results, with \textsc{r}o\textsc{bert}a clearly outperforming all other models by reaching 90 \% precision, 88 \% accuracy, recall, and \textsc{f}-score. In contrast, general-purpose \textsc{llm}s return results with accuracy close to 60 \% and around 50 \% for the remaining metrics. The confusion matrices (Figure \ref{fig:retrain_vs_llms}) reveal that these models make many mistakes between close emotions (e.g., joy/love and sadness/anger) and even confuse opposite emotions (joy/sadness).
	
	This highlights the advantage in this case of fine-tuned transformer models, which can distinguish fine-grained emotional categories thanks to their ability to capture deep contextual dependencies. Among them, \textsc{r}o\textsc{bert}a achieves the best overall results due to several key improvements over \textsc{bert}. It removes the next-sentence prediction objective, uses dynamic masking during pretraining, and is trained on a significantly larger corpus. These modifications allow \textsc{r}o\textsc{bert}a to learn richer contextual representations and generalize better to unseen examples, which is crucial for separating semantically similar emotions. In contrast, general-purpose models without task-specific training struggle to separate the categories. These values may be the result of two possible factors. The issue may stem from the use of an inappropriate prompt or the existence of similar emotional categories that are difficult to classify. Both approaches are explored below.
	
	\begin{table}[!htbp]
		\centering
		\caption{Emotion recognition results with six classes.}
		\begin{tabular}{lcccc}
			\toprule
			\bf Model & \bf Accuracy & \bf Recall & \bf Precision & \bf \textsc{f}-score \\ \midrule
			Gemma & 59.94 & 54.36 & 53.23 & 52.09 \\ 
			\textsc{gpt}-3.5 & 59.61 & 52.25 & 51.63 & 51.60 \\ 
			\textsc{ll}a\textsc{ma}-3 & 56.62 & 53.68 & 51.37 & 50.22 \\ 
			\textsc{bert} & 82.26 & 71.55 & 83.43 & 75.55 \\ 
			spaCy & 80.30 & 73.46 & 79.62 & 75.07 \\ 
			\textsc{r}o\textsc{bert}a & \bf 88.00 & \bf 88.00 & \bf 90.00 & \bf 88.00\\
			
			\bottomrule
		\end{tabular}
		\label{tab:comparison}
	\end{table}
	
	Table \ref{tab:prompt_strategy} shows the variability in performance depending on the prompt engineering strategy used. Prompt 1 presents the best \textsc{f}-score for all models, while more complex formulations, such as inverse emotion (\textit{i.e.}, strategy 5), lead to significantly lower performance, especially in Gemma and \textsc{gpt}-3.5. 
	
	The lower results are obtained with prompts 2 (binary mask) and 5 (inverse emotion). The results with the inverse emotion prompt indicate that the models cannot establish complex relationships, such as detecting an opposite emotion. In this respect, \textsc{ll}a\textsc{ma}-3 is the most powerful model, with values of accuracy \SIrange{10}{25}{\percent} above those of the others, as can be seen in Table \ref{tab:prompt_strategy} with the fifth prompt engineering strategy. The results obtained with the binary mask prompt are limited. In this case, the model trained solely on text was asked to establish a relationship between an emotion and a binary mask representation of that emotion.
	
	Furthermore, the results of prompt 2 (binary mask) show the limited ability of the models to understand the translation to a binary space. However, the numerical interpretation (prompt 4), although it does not improve the results, does offer an acceptable translation compared to the basic prompt, except in the case of Gemma, which again drastically reduces its performance.
	
	This suggests that \textsc{llm}s are highly sensitive to prompt design and that complex reasoning, such as emotion inversion, is subject to improvement. The latter emphasizes the importance of clear and concise communication for achieving effective emotion detection in zero-shot scenarios. Furthermore, the limited ability of the models to address categorization tasks in highly granular contexts, where emotions with very similar semantic meanings exist, is identified.
	
	\begin{table*}[!htbp]
		\centering
		\caption{\label{tab:prompt_strategy}Prompt engineering strategies for general-purpose \textsc{llm} models.}
		\begin{tabular}{cccccc}
			\toprule
			\bf Prompt strategy & \bf Model & \bf Accuracy & \bf Recall & \bf Precision & \bf F-score\\
			\midrule
			
			\multirow{3}{*}{1} 
			& Gemma & \bf 59.94 & \bf 54.36 & \bf 53.23 & \bf 52.09 \\ 
			& \textsc{gpt}-3.5 & 59.61 & 52.25 & 51.63 & 51.60 \\ 
			& \textsc{ll}a\textsc{ma}-3 & 56.62 & 53.68 & 51.37 & 50.22 \\
			\midrule
			
			\multirow{3}{*}{2} 
			& Gemma & 12.75 & 15.35 & 14.17 & 6.97 \\ 
			& \textsc{gpt}-3.5 & 12.12 & 13.95 & 15.35 & 10.84 \\ 
			& \textsc{ll}a\textsc{ma}-3 & 7.75 & 12.16 & 10.74 & 7.12 \\
			\midrule
			
			\multirow{3}{*}{3} 
			& Gemma & 50.62 & 42.45 & 49.28 & 41.77 \\ 
			& \textsc{gpt}-3.5 & 50.38 & 42.20 & 42.22 & 41.63 \\ 
			& \textsc{ll}a\textsc{ma}-3 & 58.0 & 49.57 & 49.62 & 49.08\\
			\midrule
			
			\multirow{3}{*}{4} 
			& Gemma & 18.44 & 26.86 & 23.38 & 16.26 \\ 
			& \textsc{gpt}-3.5 & 52.12 & 50.78 & 46.93 & 46.89 \\ 
			& \textsc{ll}a\textsc{ma}-3 & 56.81 & 53.7 & 52.05 & 51.36 \\
			\midrule
			
			\multirow{3}{*}{5} 
			& Gemma & 6.69 & 6.11 & 6.82 & 5.81 \\ 
			& \textsc{gpt}-3.5 & 23.94 & 18.51 & 18.15 & 16.46 \\ 
			& \textsc{ll}a\textsc{ma}-3 & 32.88 & 35.11 & 42.18 & 27.47 \\
			
			\bottomrule
		\end{tabular}
	\end{table*}

\begin{figure}[!htbp]
		\centering
		\subfloat[\centering Gemma]
		{{\includegraphics[width=6cm]{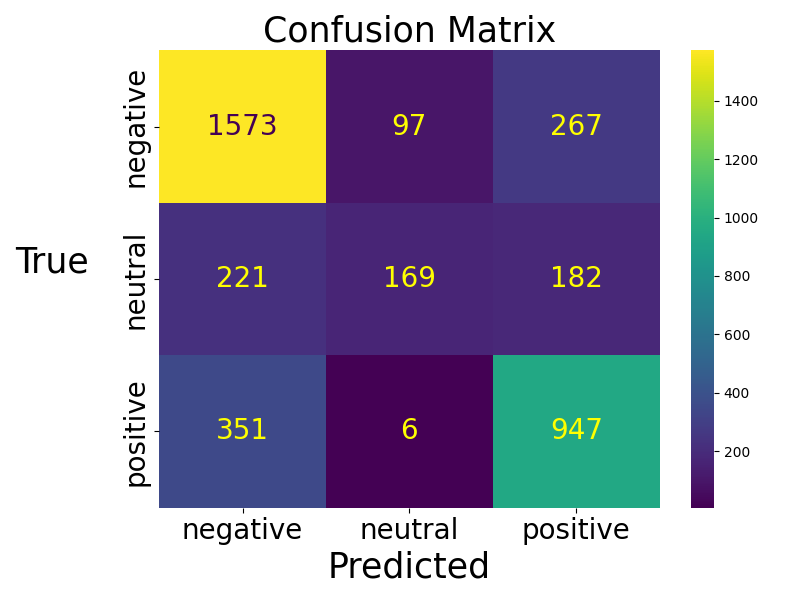}}} 
		\,
		\subfloat[\centering \textsc{gpt}-3.5]
		{{\includegraphics[width=6cm]{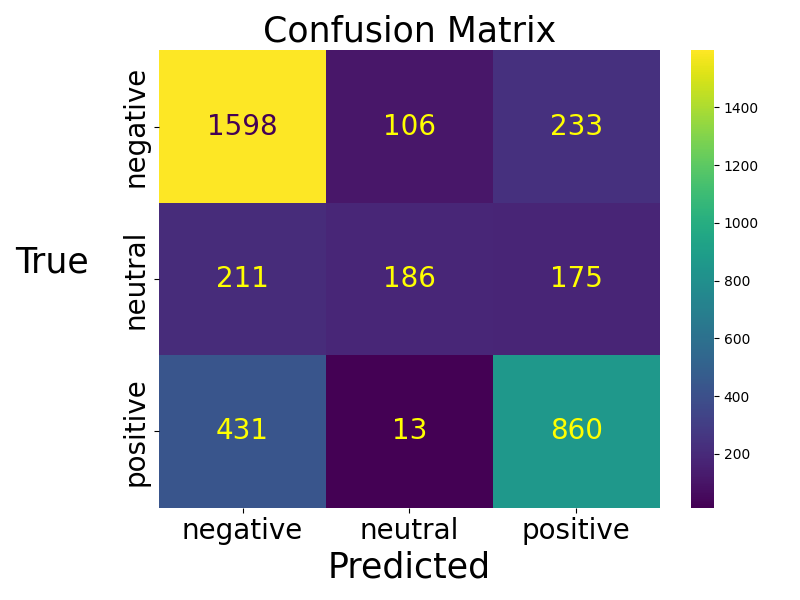}}} 
		\,
		\subfloat[\centering \textsc{ll}a\textsc{ma}-3]
		{{\includegraphics[width=6cm]{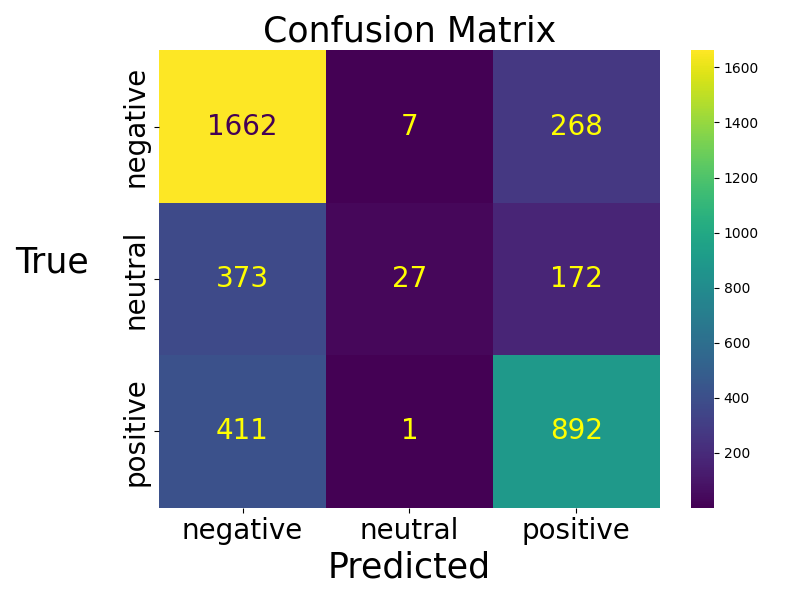}}} 
		\caption{Confusion matrix with three classes.}
		\label{fig:matrix3}
	\end{figure}

	Given that \textsc{llm}s detect emotions at the word level instead of the sentence level, the six emotion classes were grouped into three classes, applying the prompt 1 strategy: positive (love), negative (fear), and neutral (surprise). Table \ref{tab:analysis3} and Figure \ref{fig:matrix3} show the results of this approach. As can be seen in Table \ref{tab:analysis3}, the results show a notable improvement when reducing the number of emotion classes from six to three. \textsc{f}-score values increase 10 \% points compared to the previous scenario, confirming that emotion grouping reduces ambiguity and confusion between semantically close classes (\textit{e.g.}, joy and love). This result supports our claim that generalist \textsc{llm}s have difficulty discriminating fine-grained emotion categories if they have not been previously fine-tuned.
	
	The confusions decrease, reaching \SI{60}{\percent} in all metrics with Gemma and \textsc{gtp}-3.5. However, it is still far from the \SI{90}{\percent} of the fine-tuned \textsc{r}o\textsc{bert}a model with the six emotions.
	
	Additionally, the confusion matrices reflect a large number of errors committed by the neutral category toward positive and negative emotions. This effect translates into metrics that fail to exceed 70 \%, hampered by the results obtained in this intermediate category.
	
	\begin{table}[!htbp]
		\centering
		\footnotesize 
		\caption{General-purpose \textsc{llm} models with three classes.}
		\begin{tabular}{lcccc}
			\toprule
			\bf Model & \bf Accuracy & \bf Recall & \bf Precision & \bf F-score \\ \midrule
			Gemma & 66.75 & \bf 61.32 & 60.47 & 60.81 \\ 
			\textsc{gpt}-3.5 & \bf 69.34 & 60.32 & 66.72 & \bf 61.94 \\ 
			\textsc{ll}a\textsc{ma}-3 & 67.69 & 52.98 & \bf 70.69 & 50.80 \\ \bottomrule
		\end{tabular}
		\label{tab:analysis3}
	\end{table}
	
\begin{figure}[!htbp]
		\centering
		\subfloat[\centering Gemma]
		{{\includegraphics[width=6cm]{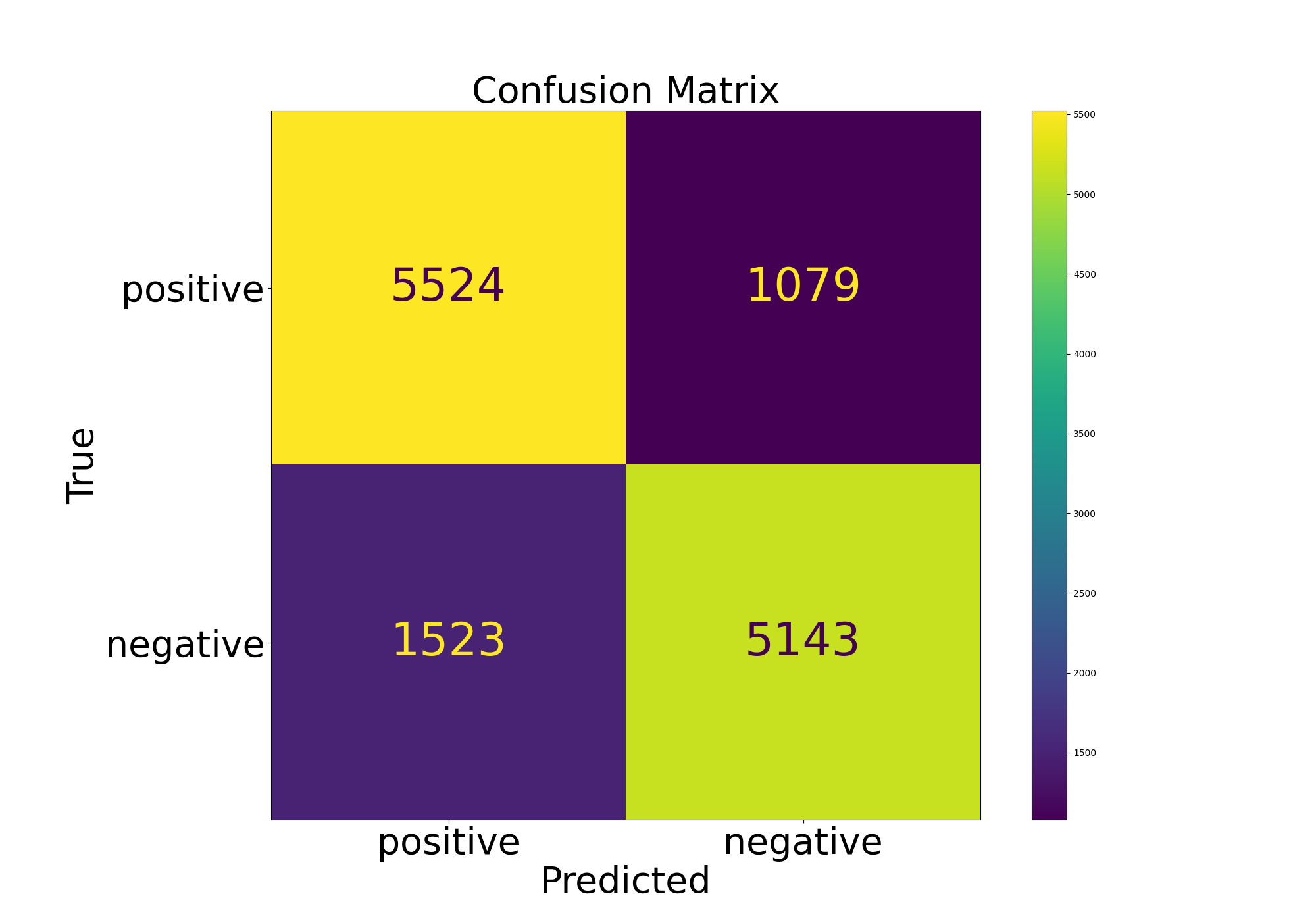}}} 
		\,
		\subfloat[\centering \textsc{gpt}-3.5]
		{{\includegraphics[width=6cm]{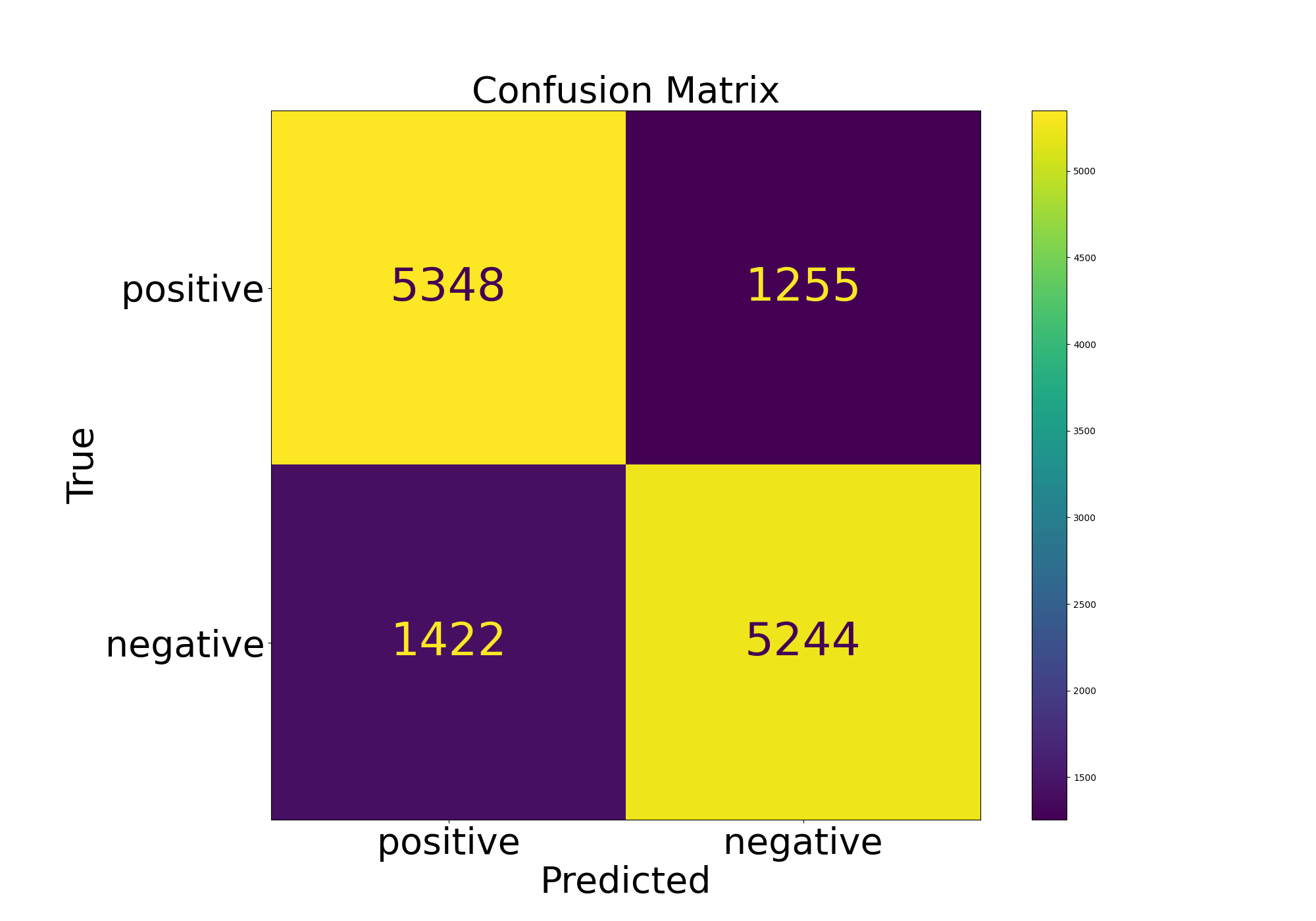}}} 
		\,
		\subfloat[\centering \textsc{ll}a\textsc{ma}-3]
		{{\includegraphics[width=6cm]{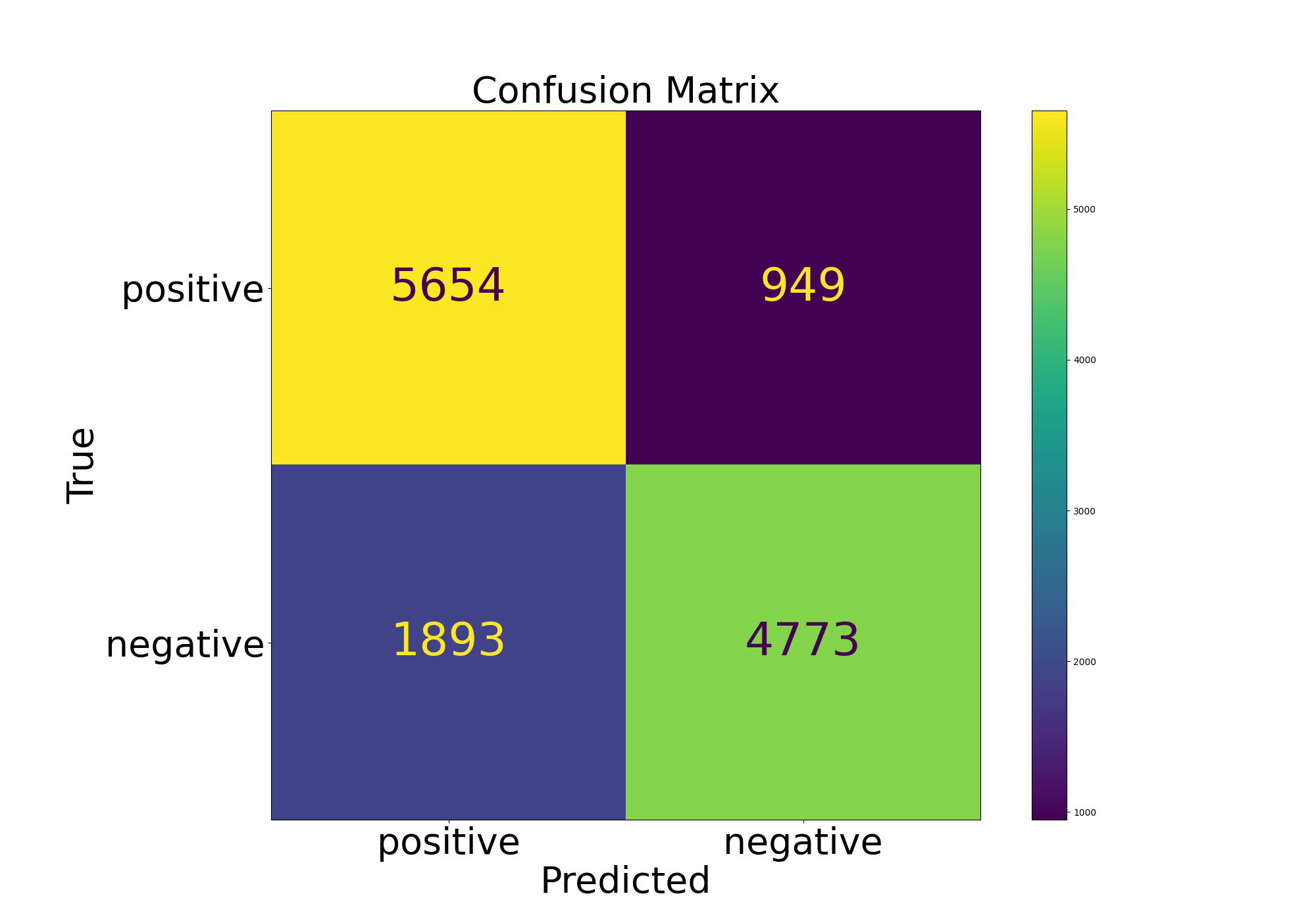}}} 
		\caption{Confusion matrix with two classes.}
		\label{fig:matrix2}
		
	\end{figure}

	Finally, Table \ref{tab:analysis2} and Figure \ref{fig:matrix2} elaborate on the binary scheme (positive \textit{versus} negative emotions). As observed in Table \ref{tab:analysis2}, the models achieve accuracies and \textsc{f}-score values greater than 78 \% in all cases. These results demonstrate that \textsc{llm}s can be effective in simplified emotional analysis tasks, even without specialized training, validating their applicability in contexts where it is sufficient to identify the general emotional polarity of the text.
	
	Furthermore, the experimental results support the conclusion that general-purpose \textsc{llm} instances have difficulty detecting more than two classes and that optimal performance requires fine-tuning.
	
	\begin{table}[!htbp]
		\centering
		\footnotesize 
		\caption{General-purpose \textsc{llm} models with two classes.}
		\begin{tabular}{lcccc}
			\toprule
			\bf Model & \bf Accuracy & \bf Recall & \bf Precision & \bf F-score \\ \midrule
			Gemma & \bf 80.39 & \bf 80.41 & \bf 80.52 & \bf 80.37 \\ 
			\textsc{gpt}-3.5 & 79.83 & 79.83 & 79.84 & 79.82 \\ 
			\textsc{ll}a\textsc{ma}-3 & 78.58 & 78.61 & 79.17 & 78.49 \\ \bottomrule
		\end{tabular}
		\label{tab:analysis2}
	\end{table}
	
	\subsection{Discussion}
	\label{sec:discussion}
	
	\cite{Amirizaniani2024} examined the extent to which \textsc{llm} understand and integrate human intentions and emotions in their open-ended answers. The approach consists of submitting human prompts from an online discussion forum and collecting and evaluating the \textsc{llm}-generated responses. The evaluation relies on (\textit{i}) humans to determine the reasoning quality of the \textsc{llm} responses; (\textit{ii}) statistical significance to establish the emotional dissimilarity between human and \textsc{llm} responses; and (\textit{iii}) metrics to quantify the semantic similarity and lexical overlap between human and \textsc{llm} responses. Although with a different objective, this work on the human-like reasoning capabilities of \textsc{llm}s addresses open-ended questions and identifies emotions and sentiments. Moreover, two of the general-purpose \textsc{llm} used are related: \textsc{ll}a\textsc{ma}-2 and \textsc{gpt}-4 versus \textsc{ll}a\textsc{ma}-3 and \textsc{gpt}-3.5, respectively. Regrettably, the emotion recognition metrics are not comparable. Nonetheless, existing similarities allow the comparison of qualitative results regarding emotion recognition with the same-family models. Best results were obtained with \textsc{gpt}-4 followed by \textsc{ll}a\textsc{ma}-2 in the case of \cite{Amirizaniani2024}. The identical behavior of the same-family models in the current proposal supports the findings.
	
	\cite{Pajon2024} explored contextual information to improve emotion recognition with general-purpose \textsc{llm}. The designed prompt always contains full or partial contextual data (the complete text or, alternatively, the emotions and polarities detected so far). The evaluation involved three public datasets and \textsc{gpt}-3.5. The Conversations and TED talks, manually labeled by the authors, achieved an \textsc{f}-score above \SI{70}{\percent} for emotions and \SI{78}{\percent} for polarities. The Short phrases, pre-labeled by default with positive and negative polarities, reached \SI{62}{\percent} with emotions and \SI{87}{\percent} with polarities. Since the data sets differ from those adopted in the current work, the results are not directly comparable. In the current work, with the emotion data set, \textsc{gpt}-3.5 achieved values of F1 around \SI{52}{\percent} (Table \ref{tab:comparison}), \SI{62}{\percent} (Table \ref{tab:analysis3}) and \SI{80}{\percent} (Table \ref{tab:analysis2}) in the detection of six classes, three classes, and two polarities, respectively. While the polarity results obtained with the emotion data set are aligned with those obtained by \cite{Pajon2024} with the Short phrases, the emotion recognition values are considerably lower. This difference may result from the impact of contextual data and the manual labeling of conversations and TED talks. 
	
	The basic prompt templates are tailored to each \textsc{llm}, resulting in three prompt templates. These templates are more detailed than the one proposed by \cite{Peng2024} and less detailed than the one adopted by \cite{Amirizaniani2024}. The size of this more extended template is related to the specific context and the dimensions of the instructions. While all templates provide the task instructions, the list of emotions, and the sentence to be examined, the \textsc{llm} prompt templates designed for this work and by \cite{Amirizaniani2024} also provide some context. However, it is far from the total or partial context provided by \cite{Pajon2024}. 
	While supplying context makes perfect sense, primarily when the data are organized by conversation, topic, or document, it has little impact when the data are made of unrelated short sentences.
	
	The best overall results regarding the detection of six emotions were achieved with the fine-tuned pre-trained \textsc{r}o\textsc{bert}a model, with all metrics above \SI{88}{\percent}.
	
	Table \ref{tab:benchmarking} presents a comparison with the most related competing work in the state of the art to validate our proposal further. As can be observed, our solution is the one that attains better performance in all metrics, with difference values of 35.63 \% in accuracy compared to \cite{Pico2024} and 33.55 \% in \textsc{f}-measure compared to \cite{Venkatesh2024}.
	
	Regarding the most recent work, our proposal continues to be the one with better performance. Compared to the works by \cite{zhang2025dialoguellm} and \cite{hong2025aer}, the differences surpass the 20 \% while being even higher between our proposal and that by \cite{bo2025toward}, in which we attained an accuracy more than 35 \% points superior.
	
	\begin{table}[htbp]
		\centering
		\small
		\caption{\label{tab:benchmarking}Benchmarking with existing research.}
		\begin{tabular}{ccccccccS[table-format=3.2]}
			\toprule
			\bf {Authorship} & \bf{Accuracy} & \bf {Recall} & \bf Precision & {\bf F-score}\\
			\midrule
			\cite{Pico2024} & 46.63 & - & 53.80 & 47.90\\
			\cite{Venkatesh2024} & - & - & - & 42.00\\
			\cite{zhang2025dialoguellm} & 61.49 & - & - & 60.52\\
			\cite{hong2025aer} & 57.87 & 65.39 & - & 58.43\\
			\cite{bo2025toward} & 45.81 & 41.56 & - & -\\
			\midrule
			\bf Proposal & \bf 82.26 & \bf 71.55 & \bf 83.43 & \bf 75.55\\
			\bottomrule
		\end{tabular}
	\end{table}
	
	The results presented indicate that generalist \textsc{llm}s show limitations in fine-grained and multi-class emotion recognition tasks. However, their performance improves significantly when both the prompt design and the emotional class structure are simplified. While complex prompts reduce accuracy, simpler ones allow for better extraction of implicit knowledge from the model. Furthermore, by grouping emotions into more general classes, the models achieve competitive results without the need for fine-tuning. This empirical evidence supports the use of a hybrid strategy, which combines the specialization of trained models with the flexibility of generalist models guided by prompt engineering.
	
	Despite the promising results, it is essential to elaborate on the limitations of the proposal. First, general-purpose \textsc{llm}s may underperform in multi-class emotion recognition tasks without applying fine-tuning. Additionally, performance is highly dependent on prompt design, thereby increasing the complexity of implementation. Moreover, the experiments focused initially on six emotions, and further analyses will be required to reflect the emotional richness present in real-world applications. Context-aware prompt engineering strategies may also be appropriate.
	
	\section{Conclusions}
	\label{sec:conclusion}
	
	With the advent of transformer models, the field of emotion recognition has experienced significant advancements. However, challenges remain regarding the exploration of open-ended queries and transformer models. Consequently, this study evaluates the performance of a large set of well-known \textsc{llm}s with different prompt strategies and emotion groupings.
	
	The evaluation stage comprises three scenarios to provide a comprehensive analysis of \textsc{llm}s for emotion detection in open-ended queries: (\textit{i}) performance of fine-tuned pre-trained models and general-purpose \textsc{llm}s using simple prompts; (\textit{ii}) effectiveness of different emotion prompt designs; and (\textit{iii}) impact of emotion grouping techniques on the performance of \textsc{llm}s.
	
	The fine-tuned \textsc{r}o\textsc{bert}a, one of the most widely used \textsc{llm}s in the literature, was the best emotion detector with metrics above \SI{88}{\percent}. Moreover, the spaCy model optimized for emotion recognition reports an average performance equivalent to that of the fine-tuned \textsc{bert}. Conversely, the general-purpose \textsc{llm}s using prompt engineering obtained results close to \SI{50}{\percent} with six emotions. Their performance improved as the number of emotion groupings decreased, reaching values near \SI{80}{\percent} for both positive and negative polarities. The prompts with the best results are the basic prompts.
	
	Shortly, the plan is to do further research on (\textit{i}) prompt design and refinement -- define a unique user prompt template, automatically refine the submitted user prompts, and automatically translate them to the different requirements of distinct general-purpose \textsc{llm} models; (\textit{ii}) perform emotion recognition through general-purpose \textsc{llm} models -- experiment with additional general-purpose \textsc{llm} models, namely Chat\textsc{glm} and \textsc{gpt}-4o, with other publicly available benchmark data sets; and (\textit{iii}) incorporating multimodal approaches that integrate text, audio, and images for emotion recognition in more complex contexts.
	
	\bibliography{bibliography.bib}{}
	\bibliographystyle{IEEEtran}
	
	\begin{IEEEbiography}[{\includegraphics[width=1in,height=1.25in,clip,keepaspectratio]{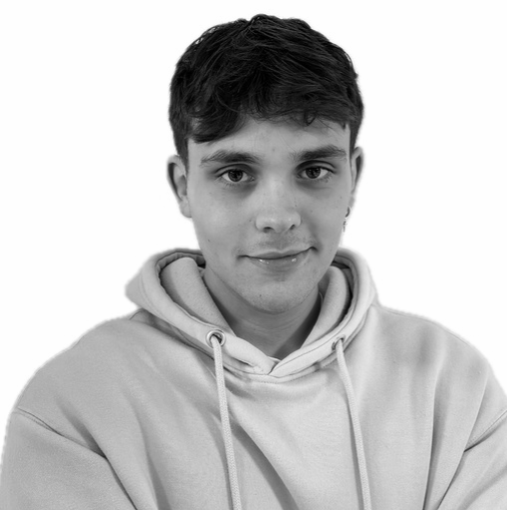}}]{Alejandro Pajón-Sanmartín} received a B.S. degree in telecommunication technologies engineering in 2023 from the University of Vigo, Spain. He is currently a researcher in the Information Technologies Group at the University of Vigo, Spain. His research interests include Natural Language Processing techniques and Large Language Models.
	\end{IEEEbiography}
	
	\begin{IEEEbiography}[{\includegraphics[width=1in,height=1.25in,clip,keepaspectratio]{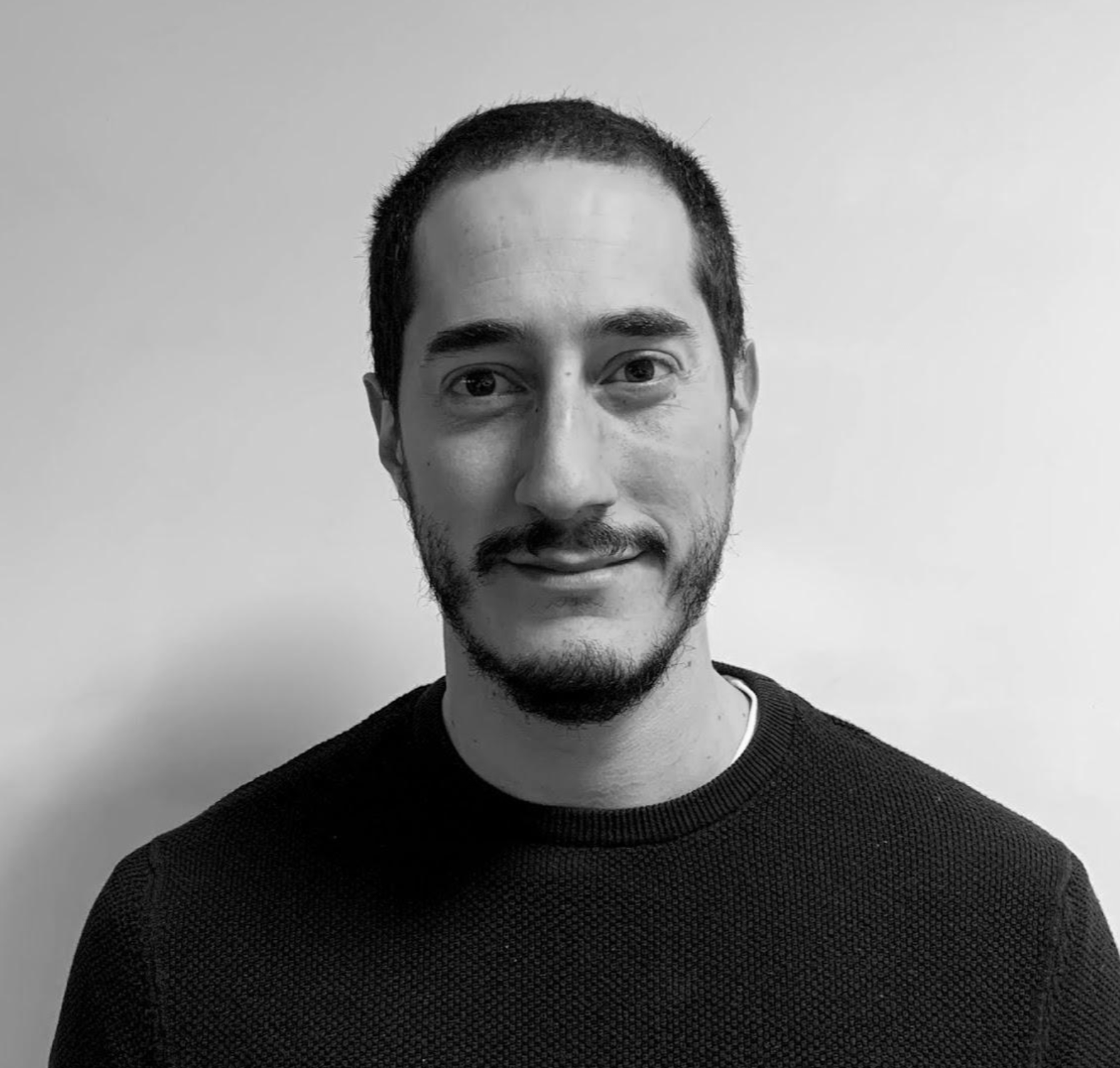}}]{Francisco de Arriba-Pérez} received a B.S. degree in telecommunication technologies engineering in 2013, an M.S. degree in telecommunication engineering in 2014, and a Ph.D. in 2019 from the University of Vigo, Spain. He is currently a researcher in the Information Technologies Group at the University of Vigo, Spain. His research encompasses the development of Machine Learning solutions for various domains, including finance and healthcare.
	\end{IEEEbiography}
	
	\begin{IEEEbiography}[{\includegraphics[width=1in,height=1.25in,clip,keepaspectratio]{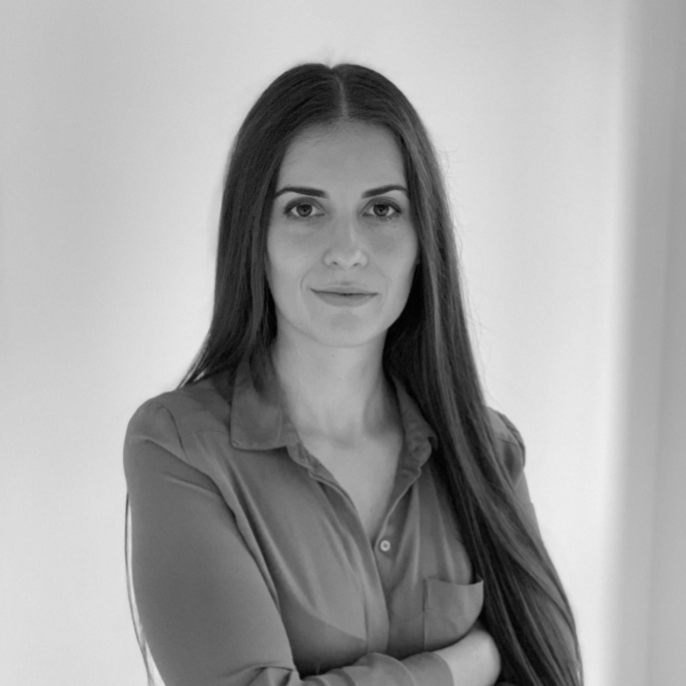}}]{Silvia García-Méndez} received a Ph.D. in Information and Communication Technologies from the University of Vigo in 2021. Since 2015, she has worked as a researcher with the Information Technologies Group at the University of Vigo. She is collaborating with foreign research centers as part of her postdoctoral stage. Her research interests include Natural Language Processing techniques and Machine Learning algorithms.
	\end{IEEEbiography}
	
	\begin{IEEEbiography}[{\includegraphics[width=1in,height=1.25in,clip,keepaspectratio]{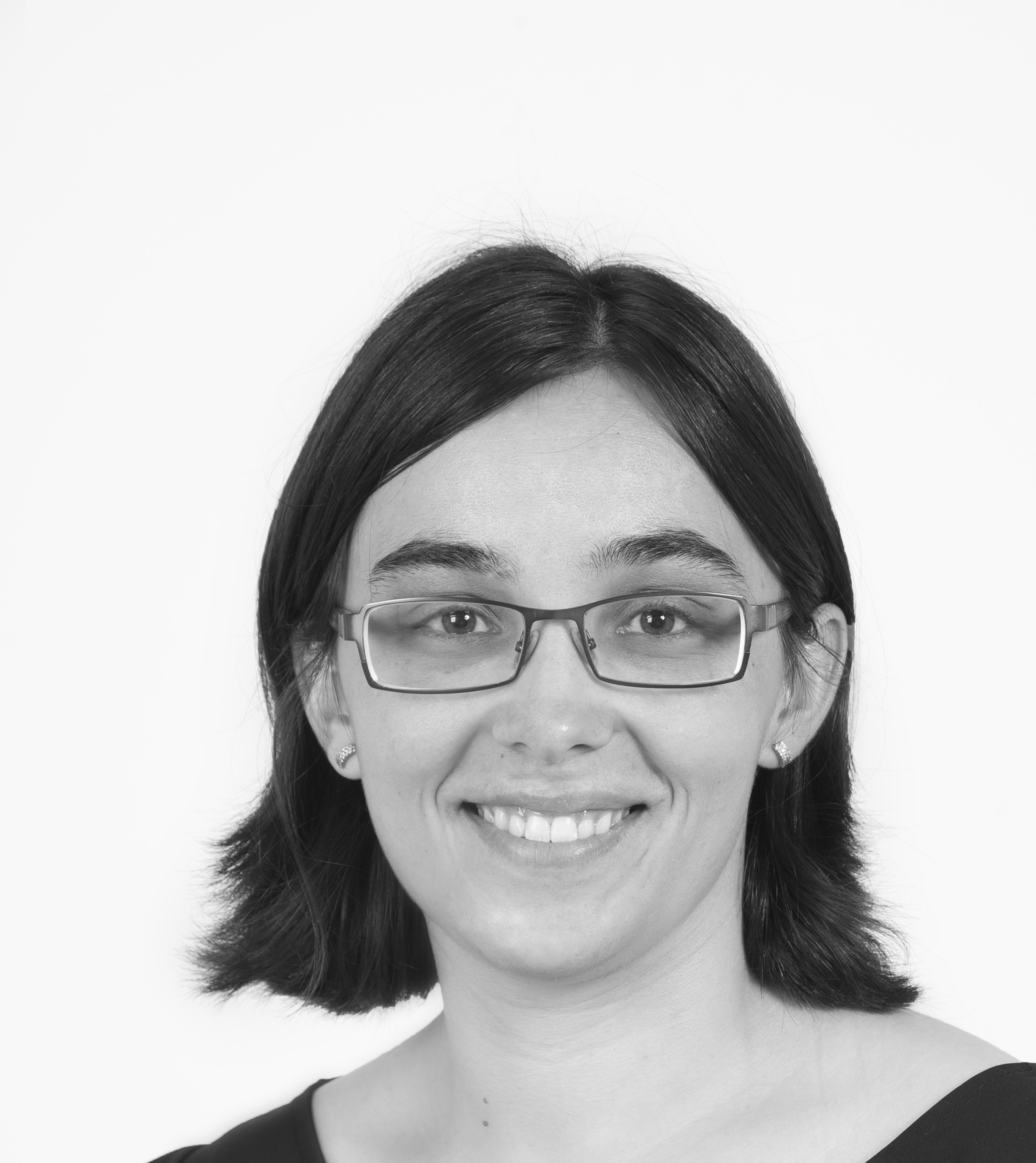}}]{Fátima Leal} holds a Ph.D. in Information
		and Communication Technologies from the University of Vigo, Spain. She is an Auxiliary Professor at Universidade Portucalense in Porto, Portugal, and a researcher at REMIT (Research on Economics, Management, and Information Technologies). Her research is based on crowdsourced information, including trust and reputation, Big Data, Data Streams, and Recommendation Systems. Recently, she has been exploring blockchain technologies for responsible data processing.
	\end{IEEEbiography}
	
	\begin{IEEEbiography}[{\includegraphics[width=1in,height=1.25in,clip,keepaspectratio]{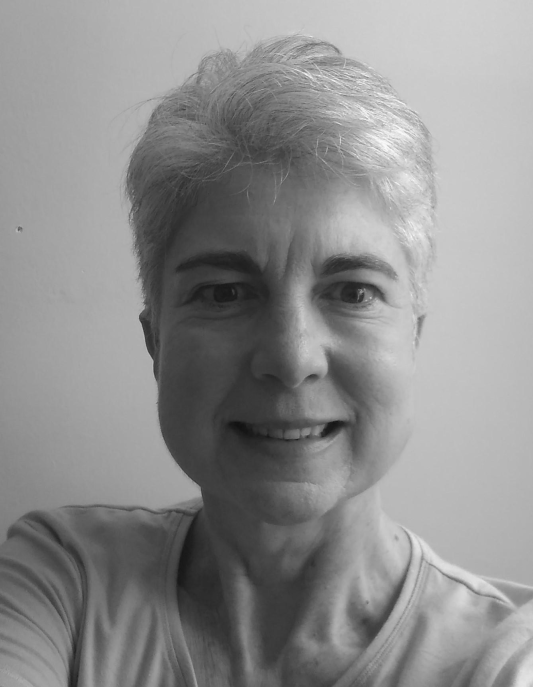}}]{Benedita Malheiro} is a Coordinator Professor
		at Instituto Superior de Engenharia do Porto, the School of Engineering of the Polytechnic of Porto, and senior researcher at \textsc{inesc} \textsc{tec}, Porto, Portugal. She holds a Ph.D., M.Sc., and a five-year degree in Electrical Engineering and Computers from the University of Porto. Her research interests include Artificial Intelligence, Computer Science, and Engineering Education. She is a member of the Association for the Advancement of Artificial Intelligence (\textsc{aaai}), the Portuguese Association for Artificial Intelligence (\textsc{appia}), the Association for Computing Machinery (\textsc{acm}), and the Professional Association of Portuguese Engineers (\textsc{oe}).
	\end{IEEEbiography}
	
	\begin{IEEEbiography}[{\includegraphics[width=1in,height=1.25in,clip,keepaspectratio]{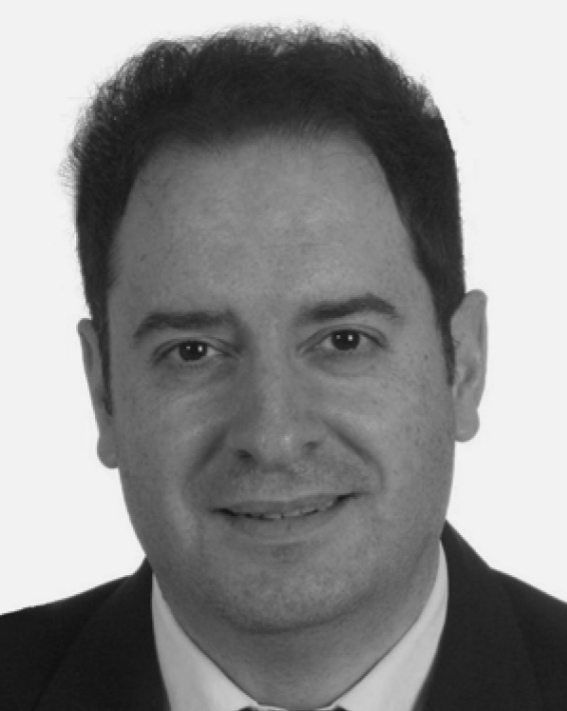}}]{Juan Carlos Burguillo-Rial} received an M.Sc. in Telecommunication Engineering and a Ph.D. in Telematics at the University of Vigo, Spain. He is currently a Full Professor at the Department of Telematic Engineering and a researcher at the AtlanTTic Research Center in Telecom Technologies at the University of Vigo. He is the area editor of the journal Simulation Modelling Practice and Theory (SIMPAT), and his topics of interest are intelligent systems, evolutionary game theory, self-organization, and complex adaptive systems.
	\end{IEEEbiography}
	
	\EOD
	
\end{document}